\pdfoutput=1
\documentclass[11pt]{article}
\usepackage{marvosym}
\usepackage[]{ACL2023}

\usepackage{times}
\usepackage{latexsym}
\usepackage{graphicx, xcolor,geometry,array}
\usepackage{amsmath}
\usepackage{booktabs}
\usepackage{multirow}
\usepackage{comment}
\usepackage{stfloats}
\usepackage{amssymb}
\usepackage{algorithm}
\usepackage{algpseudocode}

\usepackage{rotating}
\usepackage{makecell}
\usepackage{tabularx}

\usepackage[normalem]{ulem}
\useunder{\uline}{\ul}{}
\usepackage[T1]{fontenc}
\usepackage[utf8]{inputenc}

\usepackage{microtype}

\usepackage{inconsolata}

\title{Enhancing LLM Generation with Knowledge Hypergraph for Evidence-Based Medicine}
\author{
Chengfeng Dou\textsuperscript{1,2}, 
Ying Zhang\textsuperscript{3},
Zhi Jin\textsuperscript{1,2}\Letter, 
Wenpin Jiao\textsuperscript{1,2}\Letter, 
Haiyan Zhao\textsuperscript{1,2}, \\
\textbf{Yongqiang Zhao\textsuperscript{1,2}}, 
\textbf{Zhengwei Tao\textsuperscript{1,2}}\\
\textsuperscript{1} School of Computer Science, Peking University;\\ 
\textsuperscript{2} Key Laboratory of High Confidence Software Technologies(PKU), MOE, China \\
\textsuperscript{3} Beijing Key Lab of Traffic Data Analysis and Mining, Beijing Jiaotong University, Beijing, China\\
\texttt{\{chengfengdou,zhijin,jwp,zhhy.sei\}@pku.edu.cn}\\
~\texttt{\{tttzw,yongqiangzhao\}@stu.pku.edu.cn} \texttt{\{19112043\}@bjtu.edu.cn}
}

\begin{document}
\maketitle
\begin{abstract}
Evidence-based medicine (EBM) plays a crucial role in the application of large language models (LLMs) in healthcare, as it provides reliable support for medical decision-making processes. Although it benefits from current retrieval-augmented generation~(RAG) technologies, it still faces two significant challenges: the collection of dispersed evidence and the efficient organization of this evidence to support the complex queries necessary for EBM. To tackle these issues, we propose using LLMs to gather scattered evidence from multiple sources and present a knowledge hypergraph-based evidence management model to integrate these evidence while capturing intricate relationships. 
Furthermore, to better support complex queries, we have developed an Importance-Driven Evidence Prioritization (IDEP) algorithm that utilizes the LLM to generate multiple evidence features, each with an associated importance score, which are then used to rank the evidence and produce the final retrieval results.
Experimental results from six datasets demonstrate that our approach outperforms existing RAG techniques in application domains of interest to EBM, such as medical quizzing, hallucination detection, and decision support.
Testsets and the constructed knowledge graph can be accessed at \href{https://drive.google.com/file/d/1WJ9QTokK3MdkjEmwuFQxwH96j_Byawj_/view?usp=drive_link}{https://drive.google.com/rag4ebm}.
\end{abstract}
\section{Introduction}
\begin{figure}
    \centering
    \includegraphics[width=1.0\linewidth]{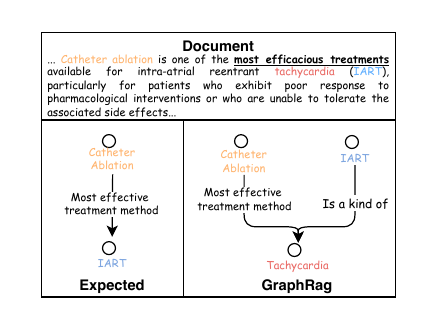}
    \caption{
    The phenomenon of mis-decomposition of complex relationships.
    LLMs omit the conditional variable ``intra-atrial reentrant'' for ``tachycardia,'' leading to incorrect extraction.
}
    \label{fig:intro}
\end{figure}
Evidence-based medicine (EBM) plays a crucial role in employing large language models~(LLMs) in the medical field, with the core principle that medical decisions must be based on objective high-quality research~\cite{greenhalgh2004effectiveness, kapoor2016types}. Although current RAG technologies can partially help in this area and effectively reduce hallucinations with LLM~\cite{ye2023cognitive,tonmoy2024comprehensive} and concerns about outdated information~\cite{gao2023retrieval}, they continue to encounter notable difficulties, such as retrieving disjointed evidence~\cite{wu2024medical} and managing intricate queries~\cite{wang2024blendfilter} necessary for EBM. 

GraphRAG~\cite{graphrag} has recently emerged to address this issue by utilizing LLMs to organize documents into a knowledge graph, thus unifying fragmented information from various documents. Using the graph retrieval algorithm, GraphRAG can access a wider range of knowledge~\cite{peng2024graph, procko2024graph}, thus improve its ability to answer complex questions. 

However, traditional knowledge graphs struggle to capture multivariate relationships~\cite{fatemi2019knowledge, masmoudi2021knowledge}, complicating the documentation of intricate relationships with several conditional variables, particularly in healthcare. To address this, LLMs must convert these into multihop paths. The retrieval process then demands an extra effort to reconstruct the original complex relationships~\cite{chen2024planongraphselfcorrectingadaptiveplanning}. Alarmingly, as demonstrated in Figure.~\ref{fig:intro}, it's tend to mis-decompose relationships when constructing knowledge graphs using LLMs. This issue significantly impedes the accurate restoration of original relationships during retrieval.

One way to tackle the aforementioned issues is reducing the dependence on relational triples when extracting knowledge and opting to depict intricate relationships with concise, unstructured text descriptions. This strategy may help to avoid excessive decomposition of knowledge units. However, it presents another challenge: without relational triples, it becomes difficult to depict associations between knowledge via a graph. 

To enhance knowledge connectivity, we propose a hypergraph-based model that incorporates entities and hierarchical hyper-relationships. Lower-tier hyper-relationships represent individual units of medical evidence, while higher-tier hyper-relationships, referred to as topics, aggregate evidence from multiple documents. 

% During the evidence retrieval process, we utilize a random walk algorithm to identify relevant topics within the hypergraph, guided by the keywords in the user's query.

For evidence retrieval on knowledge hypergraphs, we introduce the Importance-Driven Evidence Prioritization (IDEP) algorithm to rank evidence within identified topics. The algorithm begins with a random walk to identify relevant topics within the hypergraph, guided by the keywords from the user's query. IDEP then employs a LLM to extract and summarize evidence features associated with these topics, considering additional query conditions. These scores are used to compute a weighted score for each piece of evidence in the topics, which facilitates both prioritization and serves as a reference for subsequent generation tasks.

We have constructed a knowledge hypergraph that originates from 41,504 medical guidelines and drug descriptions, comprising 217,236 nodes, 433,611 topics, and 806,495 pieces of evidence. 
Our evaluations on six data sets demonstrate that our method outperforms VectorRAG~\cite{lewis2021rag} and GraphRAG~\cite{graphrag} in tasks such as medical quizzing, hallucination detection, and medical decision evaluation. 

Our contributions are as follows:
\begin{itemize}
\item We present a system for EBM tasks that constructs a knowledge hypergraph and uses RAG algorithms for importance-driven evidence prioritization, efficiently handling complex user queries and searching medical evidence.
\item We conduct a extensive experimental analyse to optimize the application of RAG in EBM. These studies aim to improve response quality, reduce inference costs, and ensure generalizability across various model scales.
\item We open-source a large-scale medical knowledge hypergraph, along with evaluation benchmarks, to facilitate future research and development in the application of RAG technology to evidence-based medicine.
\end{itemize}

\section{Related Works}
% TODO: 这个太长了，需要改写一下写的方式
% 期望达到的目标：
%   1. 说明从哪几个方面以及为什么从这些方面来介绍相关内容
%   2. 工作需要有分类
% 其他人的意见：不用把每个工作单独拿出来说，得抽象一层级，这样既可以减少篇幅，调理还会更加清楚
% This section presents a concise overview of related work in RAG and knowledge graph-based RAG, highlighting their background, benefits, and challenges essential to introduce our method.
% 用一句话说明这两块内容和标题或者研究内容的关系，点明主旨。

% \subsection{Research on Evidence-based Medicine}
% 侧重LLM，AI 应用在EBM任务上的研究发展，划定一个合适的范围
% 梳理当前研究进展
% 目前存在的棘手挑战或者尚未解决的问题（能够成为未来研究工作的那种，总结性的大方向）
% 引出我们的工作，是针对该问题的首次尝试解决或者探究，为后续工作奠定基础。

% \subsection{Retrieval-Augmented Generation}
% RAG has been shown to enhance model capabilities, offering significant benefits in various applications~\cite{fan2024survey,wang2024searching}. 
% RAG addresses language model hallucinations by integrating current and accurate data from external knowledge sources~\cite{ram2023context,gao2022precise}. 
%  前两句合并成一句即可，或者直接去掉，因为上一段已经提及了
% In healthcare, RAG holds promise for knowledge assessments~\cite{wang2024biorag}, helping medical decisions~\cite{jiang2024tc}, and improving clinical records~\cite{jadhav2024maven}.
% RAG对于医学任务，特别是EBM任务的重要性。

Recently, retrieval augmented generation~(RAG) techniques have made substantial progress in improving LLM generation capabilities~\cite{fan2024survey,wang2024searching} and alleviating hallucination issues~\cite{ram2023context,gao2022precise}. Although several studies aim to improve retrieval efficiency by optimizing document segmentation~\cite{jiang2024long}, embedding models~\cite{bge-m3}, iterative query decomposition~\cite{jiang2023active,shao2023enhancing} and refining retrieval results~\cite{yan2024corrective, wang2023learning,yu2024rank,anantha2023context}, others have concentrated on improving document management. Researches from VectorRAG~\cite{lewis2021rag} to GraphRAG~\cite{graphrag} have been explored for more efficient document representations~\cite{sarmah2024hybridrag,guo2024lightrag}, multi-hop retrieval~\cite{li2023graph,hu2024grag}, and correcting knowledge errors~\cite{xu2024knowledgeconflictsllmssurvey}.

% In recent years, retrieval-augmented generation~(RAG) techniques have made substantial progress in the medical application of LLMs~\cite{}.
% Especially, driven by its promise for knowledge assessments, RAG is widely applied to assist medical decisions~\cite{jiang2024tc}
%  and improve clinical records~\cite{jadhav2024maven}. 
 % 罗列RAG在medical方面的研究
 % 但是缺少对循证医学医学的进一步探究
 % Although the aforementioned research have explored RAG for some medical applications, evidence-based medicine remains an unresolved issue. To best of our knowledge, we 
 
% Despite its advantages, RAG faces challenges in precision and recall, which can lead to incorrect or irrelevant information selection and crucial data omission~\cite{gao2023retrieval}.
% 这句话的前半句表述是引用他人的，还是自己写的？这段要写RAG当下的研究热点在哪儿？
% To address these challenges, researchers have proposed various solutions. 
%根据下面的衔接词，这部分的研究分为五种思路？这五种思路的做法有什么不同？目前graph-RAG属于是哪一种思路。
% Some focus on minimizing missing data by improving document segmentation~\cite{jiang2024long}. 
% Others~\cite{jiang2023active,shao2023enhancing} promote iterative retrieval and generation methods to solve complex problems. 
% Research~\cite{yan2024corrective, wang2023learning} aims to filter extraneous content with a lightweight retrieval evaluator.
% Furthermore, studies~\cite{yu2024rank,anantha2023context} fine-tune language models for better document ranking to enhance RAG's quality. 
% Moreover, recent research~\cite{chirkova2024retrieval} suggests that even with top-tier retrievers and generators, task-specific prompt engineering is still crucial. % 说明提示工程很重要。

% 过渡，衔接： 从RAG到KG-RAG, 承上启下
\paragraph{Knowledge Graph RAG}
Knowledge graph-based RAG holds great promise for evidence-based medical~\cite{wu2024medical}, which employs the entity connection structure to achieve accurate and comprehensive results~\cite{peng2024graph}. 
GraphRAG uses knowledge graphs and community discovery techniques for the thorough extraction of document information, while~\citet{yang2024graphusion} introduces the Graphusion framework to improve the extraction of knowledge graphs. On this basis, 
HybridRAG~\cite{sarmah2024hybridrag} and LightRAG~\cite{guo2024lightrag} integrate graph and vector retrieval to achieve superior performance.
However, knowledge graph-based RAG encounters drawbacks in capturing complex relationships~\cite{xu2024retrieval}. 
In this work, we propose a knowledge hypergraph-based evidence management model to obtain intricate relationships of scattered evidence.

% The method of knowledge graph-based retrieval employs the entity connection structure to achieve accurate and comprehensive results~\cite{peng2024graph}. 
% 先介绍KG-RAG基本思想或者研究内容

% 引出当前存在挑战
% However, knowledge graph-based RAG faces challenges in capturing complex relationships~\cite{xu2024retrieval}. 
% 针对此挑战，现有研究工作，归纳研究思路有哪些，各自优缺点
% Recent advancements have concentrated on enhancing multihop inference capabilities. 
% To enhance multi-hop reasoning, research~\cite{li2023graph} suggests transforming knowledge graph triples into text and subsequently restructuring them using a language model. 
% GRAG~\cite{hu2024grag} utilizes a subgraph retrieval module along with a partition algorithm to achieve significant improvements in multi-hop reasoning. 
% G-Retriever~\cite{he2024gretri} frames RAG as a Prize-Collecting Steiner Tree optimization problem. 
% VectorRAG integration techniques have been extensively explored, and HybridRAG~\cite{sarmah2024hybridrag} integrates GraphRAG and VectorRAG to enhance retrieval quality. 
% LightRAG~\cite{guo2024lightrag} introduces a bilevel representation that integrates graph and vector retrieval to achieve superior performance.

% 这部分的两个工作和上文的graphRAG之间的区别？

% MedicalRAG的工作？
\paragraph{Medical RAG}
RAG has been applied in healthcare domains to assist medical decisions~\cite{jiang2024tc} and improve clinical records~\cite{jadhav2024maven}, driven by its superiority for knowledge assessment.
Medical Graph RAG~\cite{wu2024medical} employs a hierarchical approach to construct medical knowledge graphs. 
% 阐述和我们方法的关联，一方面是考虑是否延续了上述思路，另一方面是解决了前人尚未解决的问题，突出我们的贡献。
Despite their meaningful attempts, most graphs primarily describe binary relationships. Different from these studies, our approach on constructing knowledge hypergraphs is more suitable for multivariate relationships.

\section{Knowledge Hypergraph}
% 其它人的评价：看着像 Findings 而不是主会的论文。
% 主要体现在缺少形式化（问题求解）的表述，Prompt Template 过多导致看起来不学术。方法章节篇幅过长，留给实验的篇幅过短。
% TODO: 应当将 Prompt 的地方换成公式，然后把 Prompt 挪到附录里面。
Unlike graphs that depict binary relationships between pairs of entities, hypergraphs represent complex relationships among multiple entities by capturing connections within groups or clusters. Motivated by this, We introduce the knowledge hypergraph scheme for document storage in RAG to address the limitations of traditional knowledge graphs, especially in representing complex relationships in healthcare. This section outlines the definition and construction of a knowledge hypergraph scheme.

% we introduce the knowledge hypergraph scheme for document storage in RAG to address the limitations of traditional knowledge graphs in representing intricate relationships, particularly in consolidating scattered evidence with complex latent relations in healthcare. This section outlines the definition and construction of a knowledge hypergraph scheme.

\subsection{Definition of Knowledge Hypergraph}
% As shown in Figure~\ref{fig:schema}, we propose an evidence-based medical knowledge hypergraph (EbmKG) for the storage of medical documents. 
We introduce an evidence-based medical knowledge hypergraph (EbmKG) for managing medical documents, denoted as $H= \langle M,T,E; w, \mathbb{I} \rangle$. This hypergraph features medical entities $M=\{m_1,m_2,..., m_{|M|}\}$ as nodes and two types of hyperedge sets: \textit{topic} $T=\{t_1,t_2,...,t_{|T|}\}$ and \textit{evidence} $E=\{e_1,e_2,...,e_{|E|}\}$, where $|M|$, $|T|$, and $|E|$ are the counts of medical entities, topics, and evidence, respectively. The association between a medical entity $m$ and a topic $t$ is given by $w(m,t) \in [0,1]$. Evidence $e$ is included in topic $t$ if $\mathbb{I}(t, e) = 1$; otherwise, $\mathbb{I}(t, e) = 0$.

\begin{figure}[htbp]
    \centering
    \includegraphics[width=1.0\linewidth]{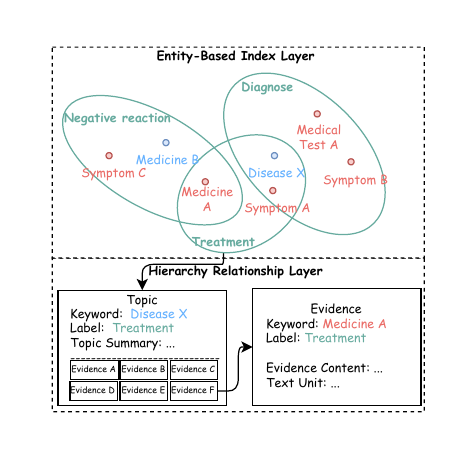}
    \caption{The Schema of EbmKG. The green ellipse denotes the hyperrelations corresponding to topic. The blue entities denote topic keywords, while red entities indicate evidence keywords. Evidence under the same topic has the same label.}
    \label{fig:schema}
\end{figure}

Figure~\ref{fig:schema} illustrates two key components: the entity-based index layer and the hierarchy relationship layer. The entity-based index layer encompasses all medical entities $M$ derived from the evidence descriptions, defining the nodes in the hypergraph. The hierarchy relationship layer, on the other hand, captures the connections among these medical entities. In this framework, evidence $e$ represents the lower-level hyperrelationship, extracted directly from medical literature. Conversely, topic $t$ is a higher-level hyperrelationship, summarizing evidence that shares common themes.

\begin{figure*}[th]
    \centering
    \includegraphics[width=1.0\linewidth]{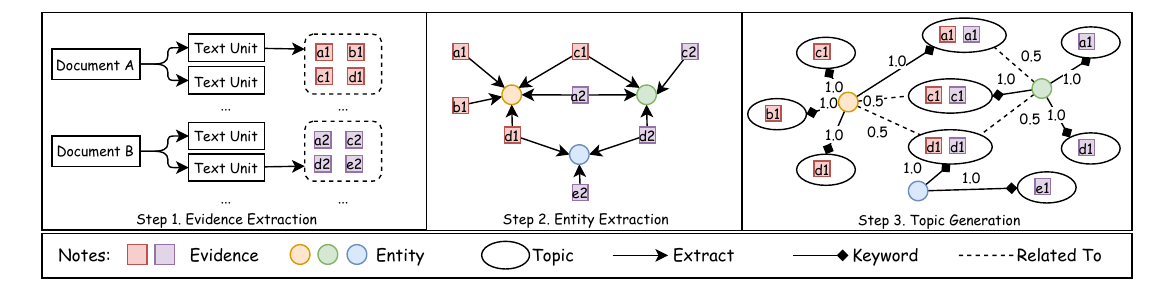}
    \caption{The construction process of EbmKG involves several key elements. Colored squares are utilized to represent evidence, with different colors distinguishing the sources of the evidence and the text within the squares indicating specific label of the evidence. Colored circles are employed to represent entities that are extracted from the evidence. White ovals represent topics, which are derived from evidence with the same aspect words.}
    \label{fig:construction}
\end{figure*}

\subsection{Graph Construction}
The construction of EbmKG proceeds through three main phases: evidence extraction, entity extraction, and topic generation, as illustrated in Figure~\ref{fig:construction}.
To help form the hypergraph, we introduce two additional concepts: keywords~($K$) and labels~($L$). Keywords stand for the core entities of hyperedges, whereas labels denote hyperedges' relationship types. A detailed explanation of each phase is provided below. All prompts we used can be find in the Appendix~\ref{sec:prompt}.

\paragraph{Document Splitting.}
Document segmentation into text units is required to facilitate efficient information extraction without exceeding the LLM context limit. A text unit is represented by $u = \langle c_\text{u}, K_\text{u} \rangle$, where $c_\text{u}$ denotes the content and $K_\text{u} \subset K$ the related keywords. We slice a document by a fixed length to obtain $c_\text{u}$, and use LLM to extract keywords from the title and abstract of the document to obtain $K_\text{u}$ and the text units in the same document split share the same $K_\text{u}$.

\paragraph{Evidence Extraction.}
We extract evidence $e = \langle c_\text{e}, l_\text{e} \rangle$ from text units $u$ using a keyword $k \in K_\text{u}$. Here, $c_\text{e}$ is the evidence description, and $l_\text{e}$ indicates the hyper-relationship type of the evidence. This extraction process can be expressed as:
\begin{gather}
c_\text{e} = \text{LLM}(k, l_\text{e}, c_\text{u}; \text{Prompt} = p_\text{e})
% l_\text{e} \in \text{Mapping}(\text{type}(k), L)
\end{gather}
where $p_\text{e}$ is the task prompt, detailed in Appendix \ref{sec:prompt} (Evidence Extraction). Since $k$ and $l_\text{e}$ together define the subject of the evidence, their combination must be meaningful. Therefore, we manually define legitimate mapping rules between keywords and hyper-relationship types, and extract evidence based only on these valid combinations. The mapping rules are provided in Appendix \ref{sec:label_def}.

\paragraph{Entity Extraction.}
After evidence collection, LLM is utilized to discern entities from the evidence employing predefined entity categories, as shown in Step 2 of Figure~\ref{fig:construction}. For a medical entity $m = \langle c_\text{m}, E_\text{m} \rangle$, where $c_\text{m}$ denotes the normalized entity name and $E_\text{m}$ is the set of relevant evidences, the extraction process is:
\begin{gather}
     f(e) = \text{LLM}(c_\text{e}, \mathbb{T}; \text{Prompt} = p_\text{m}) \\
    E_\text{m} =\{e \mid c_\text{m} \in f(e) \} 
\end{gather}
Here, $f(e) \subset M $ represents all normalized entity names extracted from evidence $e$, and $f(e)$ represents the set of entities extracted by the LLM from evidence description $c_e$ with types in $\mathbb{T}$. To merge entities from different evidence, we normalize medical terms using the Medical Subject Headings. $p_\text{m}$ are detailed in Appendix \ref{sec:prompt} Entity Extraction.

\paragraph{Topic Generation.}
In Figure~\ref{fig:construction} (Step 3), we detail the topic creation process. For each entity, we categorize related evidence by their labels. Each category forms a topic $t = \langle c_\text{t}, E_\text{t} \rangle$, where $c_\text{t}$ is the topic description, $E_t$ is the associated evidence with category $l_\text{t}$. The generation process can be formalized as:
\begin{gather}
    c_t = \text{LLM}(m, l_\text{t}, E_\text{t}; \text{Prompt}=p_\text{t}) \\
    E_t = \{e \mid e \in E_\text{m} \wedge l_\text{e} = l_\text{t}\}
\end{gather}
The prompt $p_\text{t}$ is provided in Appendix \ref{sec:prompt} under Topic Summarization.
Upon acquiring the topic, we proceed to link entities to it, thereby constructing the final hypergraph. Specifically, for each topic $t \in T$ and each medical entity $m \in M$, we begin by computing the correlation score between them using the equation~\ref{eq:weight}. If this score exceeds 0, an edge is established between the two, with the correlation score serving as the edge's weight.
\begin{equation}
    w(t, m) = \frac{|\{e \mid e \in E_\text{t} \wedge c_\text{m} \in f(e)|}{|E_\text{t}|}
    \label{eq:weight}
\end{equation}

\section{IdepRAG}
\begin{figure*}
    \centering
    \includegraphics[width=1.0\linewidth]{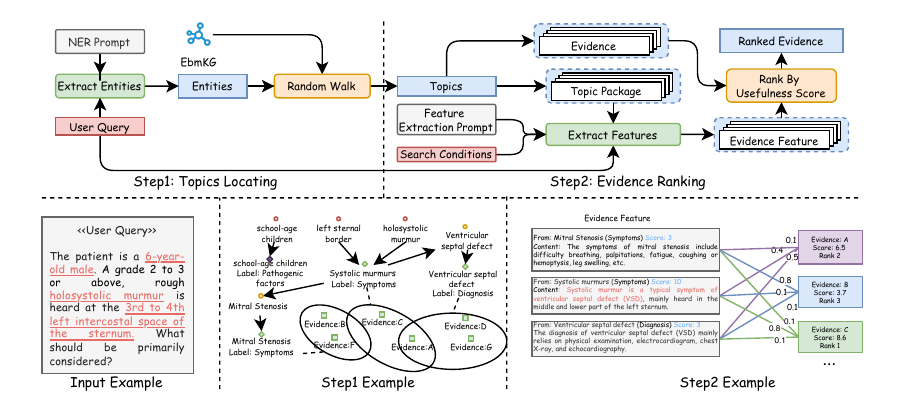}
    \caption{This figure is divided into two sections: the upper part outlines data processing, and the lower part provides an example. In 'Step 1 Example', entities in the EbmKG are shown as circles, topics as diamonds, and evidence as rectangles. Only green nodes, filtered by Personal PageRank, advance to Step 2. In 'Step 2 Example', an LLM assigns a Usefulness Score to each topic based on predefined Search Conditions, which determines the final evidence score. For NER and Feature Extraction Prompts, see Figure~\ref{fig:all_prompt} (Search Words Extraction and Evidence Features Extraction). For Search Conditions, refer to Figure~\ref{fig:score_rules}.}
    \label{fig:query}
\end{figure*}

% To handle user queries in the EBM task, it is necessary to retrieve relevant topics and evidence from EbmKG to generate the final response. Therefore, we propose an importance-driven evidence prioritization RAG approach (IdepRAG) by augmenting the LLM generation process with retrieving knowledge from EbmKG. Specifically, we design a coarse-to-fine retrieval process to acquire accurate knowledge as shown in Figure \ref{fig:query}. Firstly, we adopt the random walk algorithm to extract topics most relevant to user queries in the \textit{topic locating} stage. Then, we rank all evidence corresponding to these topics according to the usefulness score in the \textit{evidence ranking} stage. Finally, we apply the top-K ranked evidence to the generation process as the context of user inquiries for final response generation. All prompts we used in the process can be found in the Appendix~\ref{sec:prompt}.

To address user queries in the EBM task, we propose an \textbf{i}mportance-\textbf{d}riven \textbf{e}vidence \textbf{p}rioritization RAG~(IdepRAG) approach. As shown in Figure~\ref{fig:query} This method enhances LLM generation by integrating knowledge from the EbmKG. Our approach involves a two-stage retrieval process: first, a random walk algorithm identifies the most relevant topics to the query, and then, evidence related to these topics is ranked based on their usefulness scores. Finally, the top-K ranked evidence is incorporated into the generation process as contextual information for the final response. Details on the prompts used are provided in Appendix~\ref{sec:prompt}.
We will elaborate the two stages of retrieval as follows.

\subsection{Topic Locating}
This stage aims to identify potential topics relevant to the user's query inspired by community detection.
We use an LLM to extract key search terms from the query, then employ the Sentence Bert~\cite{reimers-2019-sentence-bert} to match these search terms with entities in the EbmKG. For each query term, the top $n_m$ entities with the highest similarity scores are selected, and this set of entities is denoted as $\hat{M}$.

Next, We use a random walk algorithm to identify the most relevant topics. First, we convert EbmKG into a bipartite graph $B = \langle T, M, W \rangle$, where $T$ is the set of topics, $M$ is the medical entities, and $W$ are the edge weights calculated by Equation~\ref{eq:weight}. Starting from $\hat{M} \subset M$, we perform a random walk and track the visit frequency of each topic. The $n_t$ most frequently visited topics are then selected as the output. The topic visit frequency is given by:
\begin{gather}
    P = \text{RandomWalk}(B, \hat{M}, \alpha, \beta) \\
    f(t) = \sum_i^{|P|} \mathbb{I}(t \in p_i), p_i \in P
\end{gather}
where $\alpha$ and $\beta$ represent the maximum distance of each walk and the number of iterations, respectively. $f(t)$ indicates the frequency with which topic $t$ is accessed, and $p_i$ denotes the path obtained at the i-th iteration. The random walk algorithm is capable of identifying local communities, thereby filtering out irrelevant topics, such as "school-age-children" in Figure~\ref{fig:query}.

\subsection{Evidence Ranking}
% The PPR algorithm is effective in filtering out clearly irrelevant topics from a community perspective; however, a more detailed filtration of the retained topics and their associated evidence is necessary.
After locating significant topics from EbmKG, we design a novel ranking strategy to filter the retained topics and their associated evidence.
We randomly shuffle and divide the topics into $n_p$ packages, enabling simultaneous processing by LLMs. The LLMs then extract the most relevant aspects for user queries from each package, based on predefined search conditions (Figure~\ref{fig:score_rules}).  These aspects, denoting evidence features $\mathcal{F}$, are used to compute the multi-dimensional usefulness score. In contrast to using LLMs to determine the usefulness scores directly without any additional constraints, our strategy can fully align with human preferences. 
As illustrated in "Step2 Example" of Figure~\ref{fig:query}, the usefulness scores $u_i$ for each evidence feature $f_i$ are assigned by an LLM.
For each evidence $e$, we calculate its attention scores on these evidence features as follows:  
\begin{gather}
\text{Score}(e) = \sum_i^{|\mathcal{F}|} u_i\frac{\exp(\cos(\mathbf{e}, \mathbf{f}_i))}{\sum_j^\mathcal{F} \exp(\cos(\mathbf{e}, \mathbf{f}_j))} \label{eq:weightsum} \\
f_i, u_i = \text{LLM}(q, \text{TP}_i, \text{sc}, \text{Prompt}=p_\text{r})
\label{eq:sim}
\end{gather}
where $\mathbf{e}$ and $\mathbf{f}_i$ denote the Sentence-Bert encoding of evidence $e$ and evidence features $f_i$, respectively. $u_i$ denotes the usefulness score of evidence feature $f_i$. $q$ denotes the user query, $\text{TP}_i$ denotes the i-th topic package, and $\text{sc}$ and $p_\text{r}$ are the search condition and prompt, as shown in Figures~\ref{fig:score_rules}  and~\ref{fig:all_prompt}. We rank the evidence by their usefulness scores and select the top-K to enhance LLM generation.
% However, this method has two limitations. First, the definition of usefulness can vary across different tasks, and relying on LLMs to determine usefulness without additional constraints may not fully align with human preferences. Second, when the number of topics is large, calculating the usefulness scores for each topic can be computationally expensive. 
%To address these limitations, we propose an improved evidence ranking scheme called IDEP, as illustrated in "Step2: Evidence Ranking " of Figure~\ref{fig:query}.

% In IDEP, we first concatenate the topics into a single document and divide it into chunks, forming topic chunks. This reduces the number of LLM calls required. Next, we use LLMs to extract the most useful portions for answering user queries from each topic chunk, based on user-defined search conditions. These portions, referred to as evidence features, are assigned corresponding usefulness scores. Finally, we replace the topics and their scores in Equation~\ref{eq:weightsum} and ~\ref{eq:sim} with the evidence features and their respective usefulness scores to calculate the final evidence scores.
\section{Experiment Setting}
\begin{table*}[htbp]
\centering
\small
\begin{tabular}{@{}lccccccc@{}}
\toprule
\multicolumn{1}{l|}{\multirow{2}{*}{\textbf{Model}}} & \multicolumn{3}{c|}{\textbf{Medical QA}}                                    & \multicolumn{2}{c|}{\textbf{\begin{tabular}[c]{@{}c@{}}Hallucination Detection\end{tabular}}} & \multicolumn{1}{c|}{\textbf{\begin{tabular}[c]{@{}c@{}}Decision Support\end{tabular}}} & \multicolumn{1}{l}{\multirow{2}{*}{\textbf{Avg}}} \\ \cmidrule(lr){2-7}
\multicolumn{1}{l|}{}                                & \textbf{MedQA-ML} & \textbf{NLPEC} & \multicolumn{1}{c|}{\textbf{CMB-Clin}} & \textbf{CMHE-D}                      & \multicolumn{1}{c|}{\textbf{CMHE-G}}                      & \multicolumn{1}{c|}{\textbf{DDA}}                                                            & \multicolumn{1}{l}{}                              \\ \midrule
\multicolumn{8}{c}{\textit{Qwen2.5-72B-Instruct}}                                                                                                                                                                                                                                                                                                                                                 \\ \midrule
\multicolumn{1}{l|}{w/o RAG}                         & \textbf{90.6}     & 90.3           & \multicolumn{1}{c|}{90.0}              & 71.4                                 & \multicolumn{1}{c|}{71.5}                                 & \multicolumn{1}{c|}{60.9}                                                                    & 79.1                                              \\
\multicolumn{1}{l|}{VectorRAG}                       & {\ul 90.4}        & \textbf{90.5}  & \multicolumn{1}{c|}{88.0}              & {\ul 71.7}                           & \multicolumn{1}{c|}{72.9}                                 & \multicolumn{1}{c|}{{\ul 63.4}}                                                              & {\ul 79.5}                                        \\
\multicolumn{1}{l|}{GraphRAG}                        & 83.4              & 82.2           & \multicolumn{1}{c|}{{\ul 90.2}}        & 70.3                                 & \multicolumn{1}{c|}{{\ul 73.4}}                           & \multicolumn{1}{c|}{58.5}                                                                    & 76.3                                              \\
\multicolumn{1}{l|}{IdepRAG}                         & 89.9              & \textbf{90.5}  & \multicolumn{1}{c|}{\textbf{90.7}}     & \textbf{72.0}                        & \multicolumn{1}{c|}{\textbf{73.5}}                        & \multicolumn{1}{c|}{\textbf{77.8}}                                                           & \textbf{82.4}                                     \\ \midrule
\multicolumn{8}{c}{\textit{GPT-4o-2024-08-06}}                                                                                                                                                                                                                                                                                                                                                    \\ \midrule
\multicolumn{1}{l|}{w/o RAG}                         & {\ul 83.2}        & 84.4           & \multicolumn{1}{c|}{82.7}              & 67.0                                 & \multicolumn{1}{c|}{75.9}                                 & \multicolumn{1}{c|}{{\ul 72.4}}                                                              & {\ul 77.6}                                        \\
\multicolumn{1}{l|}{VectorRAG}                       & 83.0              & {\ul 86.2}     & \multicolumn{1}{c|}{81.8}              & 66.7                                 & \multicolumn{1}{c|}{{\ul 77.9}}                           & \multicolumn{1}{c|}{68.3}                                                                    & 77.3                                              \\
\multicolumn{1}{l|}{GraphRAG}                        & 78.1              & 77.3           & \multicolumn{1}{c|}{{\ul 83.4}}        & {\ul 67.4}                           & \multicolumn{1}{c|}{75.6}                                 & \multicolumn{1}{c|}{63.3}                                                                    & 74.2                                              \\
\multicolumn{1}{l|}{IdepRAG}                         & \textbf{84.1}     & \textbf{87.1}  & \multicolumn{1}{c|}{\textbf{84.7}}     & \textbf{67.6}                        & \multicolumn{1}{c|}{\textbf{78.6}}                        & \multicolumn{1}{c|}{\textbf{75.9}}                                                           & \textbf{79.6}                                     \\ \bottomrule
\end{tabular}
\caption{The performance of our proposed methods compared to other RAG baselines on six datasets is presented. Bolding indicates the best result, while underlining indicates the second best result.}
\label{tab:main_result}
\end{table*}

\paragraph{Datasets}

Our approach is assessed using six benchmarks categorized into three distinct groups: (1) Medical question answering, incorporating two multiple-choice datasets—MedQA \cite{jin2020medqa} derived from medical licensure exams in Mainland China, and NLPEC \cite{li2020medical}, associated with pharmacist exams—as well as one open-ended QA dataset, CMB-clin, which is centered on analyzing patient reports. (2) Hallucination detection employing the CMHD \cite{cmedbenchmark} dataset, which is subdivided into discriminative (CMHD-D) and generative tasks (CMHD-G). (3) Decision support using the DDA dataset, which we developed specifically for intricate medical decision-making.

\paragraph{Baselines}
We compared our proposed method with two established baselines: VectorRAG~\cite{lewis2021rag} and GraphRAG~\cite{graphrag}. Our experiments utilized the Qwen2.5~\cite{qwen2.5} and GPT-4o language models. This study aims to investigate the effective utilization of unstructured text for RAG. To ensure a fair comparison, all baselines were required to use the same data sources, which included 41,504 publicly available drug descriptions and medical guidelines collected from the web. 
% For specific implementation details, please see Appendix~\ref{sec:imp_details}.

\paragraph{Metrics} 
Given the dual impact of the LLM and Retriever on the RAG system, we evaluated its performance from two aspects: retrieval and generation quality. For generation, we used accuracy for MedQA and NLPEC, F1-score for CMHD-D, and key point coverage for CMHD-G, DDA, and CMB-clin. Details on datasets and key point evaluation are in Appendices \ref{sec:data_des} and \ref{sec:kp_eval}. For retrieval, we randomly selected 200 samples from each dataset and compared IdepRAG with VectorRAG and GraphRAG using recall, precision, and a combined dominance score. 
% Metric calculations are in Appendix \ref{sec:appendix_com}.
More details on datasets, implementations, and metrics are in Appendix~\ref{sec:data_des},~\ref{sec:imp_details}, and \ref{sec:appendix_com}, respectively.

\section{Experimental Results}
In this section, we examine the effectiveness of IdepRAG from two perspectives: generation evaluation and retrieval evaluation. \textbf{For details on the ablation study, please refer to Appendix~\ref{sec:ab_study}.}

\subsection{Assessing the Quality of Generation}
Table~\ref{tab:main_result} compares IdepRAG with other RAG baselines across six datasets, demonstrating that IdepRAG outperforms the other methods. To understand these results, we will analyze each baseline in the context of specific tasks.

\begin{table*}[htbp]
\centering
\small
\begin{tabularx}{\textwidth}{@{}l l *{6}{>{\centering\arraybackslash}X} c@{}}
\toprule
\multirow{2}{*}{\textbf{Dataset}} & \multirow{2}{*}{\textbf{Model}} & \multicolumn{3}{c}{\textbf{Recall}} & \multicolumn{3}{c}{\textbf{Precision}} & \multirow{2}{*}{\textbf{Advantage}} \\
\cmidrule(lr){3-5} \cmidrule(lr){6-8}
& & \textbf{Win} & \textbf{Tie} & \textbf{Loss} & \textbf{Win} & \textbf{Tie} & \textbf{Loss} & \\
\midrule
\multirow{2}{*}{MedQA} & Idep vs. Vector & 48.5 & 17.5 & 34.0 & 69.0 & 2.5 & 28.5 & 28.4$\uparrow$ \\
& Idep vs. Graph & 41.0 & 6.5 & 52.5 & 57.0 & 1.0 & 42.0 & -0.4$\downarrow$ \\
\midrule
\multirow{2}{*}{NLPEC} & Idep vs. Vector & 69.2 & 8.3 & 22.4 & 81.2 & 3.1 & 15.6 & 58.8$\uparrow$ \\
& Idep vs. Graph & 64.2 & 7.4 & 28.4 & 73.2 & 0.0 & 26.7 & 42.4$\uparrow$ \\
\midrule
\multirow{2}{*}{CMHD} & Idep vs. Vector & 8.0 & 85.0 & 7.0 & 14.5 & 76.0 & 9.5 & 13.2$\uparrow$ \\
& Idep vs. Graph & 7.8 & 60.4 & 31.7 & 19.8 & 45.8 & 34.3 & -69.8$\downarrow$ \\
\midrule
\multirow{2}{*}{DDA} & Idep vs. Vector & 78.0 & 18.0 & 4.0 & 95.0 & 0.0 & 5.0 & 90.2$\uparrow$ \\
& Idep vs. Graph & 31.0 & 46.0 & 23.0 & 78.0 & 1.0 & 21.0 & 32.8$\uparrow$ \\
\midrule
\multirow{2}{*}{CMB-Clin} & Idep vs. Vector & 83.0 & 0.5 & 16.5 & 84.5 & 0.0 & 15.5 & 64.0$\uparrow$ \\
& Idep vs. Graph & 72.0 & 0.5 & 27.5 & 74.0 & 0.0 & 26.0 & 46.4$\uparrow$ \\
\bottomrule
\end{tabularx}
\caption{Fine-grained comparisons with IdepRAG and other RAG approaches.}
\label{tab:compare}
\end{table*}

\paragraph{Medical QA and Hallucination Detection}
VectorRAG and GraphRAG show different performance based on the task type. For closed-ended QA tasks (e.g., MedQA, NLPEC), where answers are directly in the context and require precise retrieval, VectorRAG outperforms GraphRAG. Conversely, GraphRAG excels in open-ended QA tasks (e.g., CMB Clin) due to its ability to retrieve diverse information~\cite{graphrag, guo2024lightrag}. The hallucination detection task, which involves verifying the accuracy of medical statements, further highlights these differences. This task can be considered semi-open, as it includes both closed and open aspects. For example, verifying a doctor's medication recommendations is a closed QA task because the drug name is specified. However, checking for missed diseases in a diagnosis is an open-ended QA task because the missed diseases are not explicitly mentioned. In such complex scenarios, neither VectorRAG nor GraphRAG can consistently perform well.

In contrast, IdepRAG leverages the advantages of both VectorRAG and GraphRAG. By utilizing our proposed hypergraph structure, IdepRAG effectively retrieves a diverse range of literature while minimizing errors during the retrieval process. This allows IdepRAG to maintain high performance in both closed and open QA tasks.

\paragraph{Decision Support}
This experiment uses the RAG system to find out evidence to determine if medical decisions are safe. Ensuring safety in medicine involves multiple checks, such as treatment appropriateness and drug side effect assessment. Effective evidence retrieval requires a robust medical reasoning retriever. As shown in Table~\ref{tab:main_result} (DDA), both semantic similarity-based (VectorRAG) and knowledge graph-based (GraphRAG) retrieval methods fall short in providing useful evidence. In contrast, IdepRAG, which allows for detailed search criteria (Figure~\ref{fig:score_rules} (DDA)), significantly enhances performance in specialized tasks, highlighting its flexibility and effectiveness.

% This experiment aims to employ the RAG system to identify evidence supporting the safety of medical decisions. In the medical field, ensuring safety involves multiple checks, such as the appropriateness of treatment and the assessment of drug side effects. Retrieving relevant evidence requires a retriever with robust medical reasoning capabilities. Experimental results, as shown in Table~\ref{tab:main_result} (DDA), indicate that neither semantic similarity-based retrieval (VectorRAG) nor knowledge graph-based retrieval (GraphRAG) meet the necessary standards for relevance and comprehensiveness. In contrast, IdepRAG allows for the specification of detailed search criteria, as illustrated in Figure~\ref{fig:score_rules} (DDA), thereby significantly enhancing its performance in specialized tasks. This underscores the flexibility and effectiveness of our approach.

\subsection{Assessing the Quality of Retrieval}
\label{sec:comp}
We found that LLMs' robustness can reduce the impact of low-quality references on the output. Thus, evaluating only the generated results is inadequate. This section assesses the retrieval process quality directly. The conditions under which RAG improves LLM output are discussed in Section~\ref{sec:discussion}.

The evaluation results are shown in Table~\ref{tab:compare}. These results indicate that GraphRAG generally performs better than VectorRAG in retrieval quality. IdepRAG often outperforms GraphRAG, highlighting the benefits of our system. Interestingly, while GraphRAG has lower accuracy scores on some datasets, GPT-4o scores higher for MedQA and NLPEC with GraphRAG compared to VectorRAG. Error analysis shows that these inaccuracies come from GraphRAG's difficulty in handling complex relationships, as explained in Appendix~\ref{sec:case_study}.

Furthermore, in the CHMD dataset, GraphRAG excels in comparison-based assessments but lags behind IdepRAG in keypoint-based ones. This is because IdepRAG is tailored to retrieve evidence that challenges the physician's decision (see Fig. ~\ref{fig:score_rules} CMHD), boosting its keypoint-based scores. Conversely, GraphRAG retrieves both supporting and opposing evidence, which LLMs view as more comprehensive. However, for hallucination detection, retrieving supportive evidence is less valuable, making our approach more practical.
\section{Analysis and Discussion}

\begin{figure}
    \centering
    \includegraphics[width=1.0\linewidth]{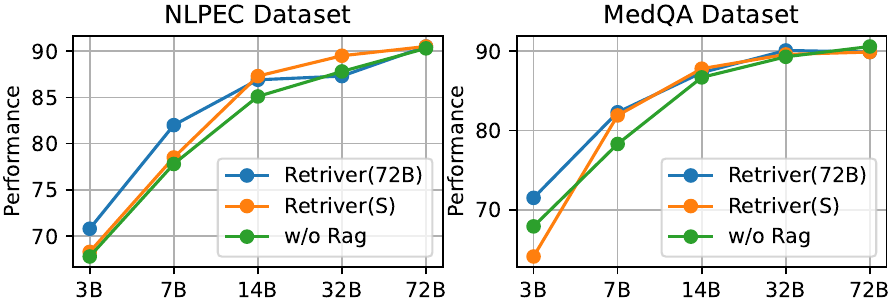}
    \caption{
    The Influence of Model Parameter Numbers on IdepRAG. \textit{Retriever (72B)}: This configuration uses a model with 72 billion parameters when executing the IDEP algorithm to retrieve high quality evidence. \textit{Retriever(S)}: This setup uses models with the same parameter scales in both the retrieval and generation phases.
    }
    \label{fig:tendence}
\end{figure}

\label{sec:discussion}
While IdepRAG outperforms VectorRAG and GraphRAG in evidence retrieval, its performance improvement is relatively modest on certain tasks despite providing more useful evidence. This phenomenon results from two most likely reasons: (1) LLMs' limited ability to effectively utilize retrieved evidence, particularly in assessing its sufficiency, leading to incorrect conclusions; and (2) IdepRAG's insufficient recall of key evidence, possibly due to incomplete document coverage in EbmKG or inherent method limitations. We will investigate each reason in the following subsections.
% These factors will be addressed in the following subsections.

\subsection{The Impact of LLM Ability}

\begin{figure}
    \centering
    \includegraphics[width=1.0\linewidth]{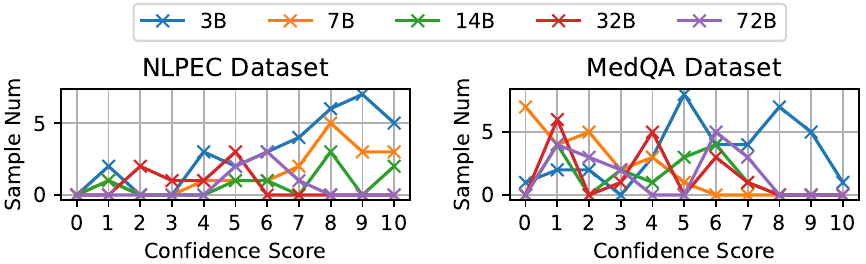}
    \caption{Confidence distributions for evidence completeness when LLMs err. Sample Num denotes the frequency of LLM errors.}
    \label{fig:frequency_distributions}
\end{figure}

It is well established that the number of parameters significantly influences LLM capabilities. To investigate this impact on evidence utilization, we conducted experiments using five Qwen2.5 models, ranging from 3 billion to 72 billion parameters, on the MedQA and NLPEC datasets (Figure~\ref{fig:tendence}). It can be found that the RAG framework substantially enhances the performance of smaller models. Models with parameters no more than 14 billion see a 4-point improvement in correctness when using high-quality evidence from a 72-billion-parameter retriever. However, this reliance on RAG diminishes as model size increases, likely because larger models have more inherent knowledge.

Interestingly, the RAG framework does not significantly enhance the performance of the 3B model to the same extent as it does for the 7B model. This finding contradicts the expectation that smaller models would benefit more from such a framework. To investigate this, we randomly selected 100 samples from MedQA for error analysis. Specifically, we had LLMs of various sizes evaluate the completeness of evidence from the Retriever(72B) on their error sample sets. The LLMs rated completeness on a scale from 0 to 10: scores >7 indicate high completeness, 3-7 indicate partial completeness, and $\le 3$ indicate low completeness. Results are shown in Figure \ref{fig:frequency_distributions}.

It can be found that larger models $(\ge 14B)$ often judged the retrieval evidence as incomplete or irrelevant when they made mistakes, while smaller models tended to rate the evidence as highly complete. This suggests that larger models are more effective at leveraging the provided evidence, explaining why RAG does not significantly improve the 3B model as much as the 7B model.

\begin{figure}
    \centering
    \includegraphics[width=1.0\linewidth]{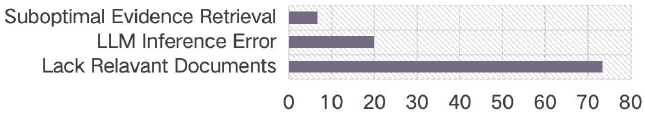}
    \caption{Distribution of error causes for IdepRAG. See Section \ref{sec:etd} for error type definitions.}
    \label{fig:error_bar}
\end{figure}

\begin{table}[htbp]
\centering
\small
\begin{tabular}{ccc}
\toprule
Model                        & MedQA & NLPEC \\ \midrule
w/o RAG & 91.0  & 90.5  \\
IdepRAG & 94.5  & 95.0  \\ \bottomrule
\end{tabular}
\caption{IdepRAG performance under high document coverage settings.}
\label{tab:complete}
\end{table}

\subsection{The Impact of Retrieval Quality}
In the previous section, we examined the key factors influencing LLMs' use of evidence. Now, we analyze how retrieval quality impacts LLM output. Figure~\ref{fig:tendence} shows that the Retriever(72B) setting outperforms the smaller Retriever(S) in accuracy, indicating lower-quality evidence retrieval by smaller models. This issue improves with model sizes over 14B, suggesting that smaller models may require additional fine-tuning for effective retrieval.

We analyzed errors in the 72B model using 50 randomly selected entries from the MedQA dataset. Approximately 73\% of the errors were due to insufficient coverage of medical literature, as our collected literature primarily focuses on clinical guidelines and lacks comprehensive coverage of national healthcare policies, biochemical principles, and other related areas. Only 6.7\% of the errors were due to poor evidence retrieval, indicating the effectiveness of our approach. To further quantify the impact of the Lack Relevant Documents, we manually verified 200 test samples from MedQA and NLPEC, ensuring answers were included in the documents. Table ~\ref{tab:complete} shows that IdepRAG significantly improves LLM scores on these samples, highlighting the importance of document coverage for RAG performance as model size increases.

\section{Conclusion}
In this work, we introduce a method for representing knowledge using hypergraphs, leveraging a LLM to build a hypergraph of medical knowledge. We also present an IdepRAG algorithm to enhance the retrieval process for intricate queries. Experimental results in six data sets show that our approach surpasses current RAG techniques in the areas of medical quizzing, hallucination detection, and decision support.

\section*{Limitations}
Evidence-based medicine (EBM) encompasses not only the retrieval of evidence but also the evaluation of its quality. In this study, the medical guidelines and drug instructions employed are of high quality, as they are based on physicians' summaries of multiple clinical cases. Consequently, the current method does not prioritize stringent control of evidence quality, which represents a primary limitation of this analysis. However, medical guidelines and drug instructions cannot comprehensively address all medical contexts, highlighting the ongoing challenge of deriving high-quality evidence from low-quality case data, and it will be a focal point of our subsequent research.

\section*{Ethics Statement}
The medical guidelines and drug instructions for constructing the knowledge hypergraph, as well as the evaluation datasets, are publicly available. All test data has been verified by professional doctors and de-identified to ensure no risk of exposing patient-sensitive information.

% \section*{Acknowledgements}

\bibliography{myacl}

\begin{thebibliography}{40}
\expandafter\ifx\csname natexlab\endcsname\relax\def\natexlab#1{#1}\fi

\bibitem[{Anantha et~al.(2023)Anantha, Bethi, Vodianik, and Chappidi}]{anantha2023context}
Raviteja Anantha, Tharun Bethi, Danil Vodianik, and Srinivas Chappidi. 2023.
\newblock Context tuning for retrieval augmented generation.
\newblock \emph{arXiv preprint arXiv:2312.05708}.

\bibitem[{Chen et~al.(2024{\natexlab{a}})Chen, Xiao, Zhang, Luo, Lian, and Liu}]{bge-m3}
Jianlv Chen, Shitao Xiao, Peitian Zhang, Kun Luo, Defu Lian, and Zheng Liu. 2024{\natexlab{a}}.
\newblock \href {http://arxiv.org/abs/2402.03216} {Bge m3-embedding: Multi-lingual, multi-functionality, multi-granularity text embeddings through self-knowledge distillation}.

\bibitem[{Chen et~al.(2024{\natexlab{b}})Chen, Tong, Jin, Sun, Ye, and Xiong}]{chen2024planongraphselfcorrectingadaptiveplanning}
Liyi Chen, Panrong Tong, Zhongming Jin, Ying Sun, Jieping Ye, and Hui Xiong. 2024{\natexlab{b}}.
\newblock \href {http://arxiv.org/abs/2410.23875} {Plan-on-graph: Self-correcting adaptive planning of large language model on knowledge graphs}.

\bibitem[{Dongfang et~al.(2020)Dongfang, Baotian, Qingcai, Weihua, and Anqi}]{li2020medical}
Li~Dongfang, Hu~Baotian, Chen Qingcai, Peng Weihua, and Wang Anqi. 2020.
\newblock Towards medical machine reading comprehension with structural knowledge and plain text.
\newblock In \emph{EMNLP}.

\bibitem[{Edge et~al.(2024)Edge, Trinh, Cheng, Bradley, Chao, Mody, Truitt, and Larson}]{graphrag}
Darren Edge, Ha~Trinh, Newman Cheng, Joshua Bradley, Alex Chao, Apurva Mody, Steven Truitt, and Jonathan Larson. 2024.
\newblock From local to global: A graph rag approach to query-focused summarization.
\newblock \emph{arXiv preprint arXiv:2404.16130}.

\bibitem[{Fan et~al.(2024)Fan, Ding, Ning, Wang, Li, Yin, Chua, and Li}]{fan2024survey}
Wenqi Fan, Yujuan Ding, Liangbo Ning, Shijie Wang, Hengyun Li, Dawei Yin, Tat-Seng Chua, and Qing Li. 2024.
\newblock A survey on rag meeting llms: Towards retrieval-augmented large language models.
\newblock In \emph{Proceedings of the 30th ACM SIGKDD Conference on Knowledge Discovery and Data Mining}, pages 6491--6501.

\bibitem[{Fatemi et~al.(2019)Fatemi, Taslakian, Vazquez, and Poole}]{fatemi2019knowledge}
Bahare Fatemi, Perouz Taslakian, David Vazquez, and David Poole. 2019.
\newblock Knowledge hypergraphs: Prediction beyond binary relations.
\newblock \emph{arXiv preprint arXiv:1906.00137}.

\bibitem[{Gao et~al.(2022)Gao, Ma, Lin, and Callan}]{gao2022precise}
Luyu Gao, Xueguang Ma, Jimmy Lin, and Jamie Callan. 2022.
\newblock Precise zero-shot dense retrieval without relevance labels.
\newblock \emph{arXiv preprint arXiv:2212.10496}.

\bibitem[{Gao et~al.(2023)Gao, Xiong, Gao, Jia, Pan, Bi, Dai, Sun, and Wang}]{gao2023retrieval}
Yunfan Gao, Yun Xiong, Xinyu Gao, Kangxiang Jia, Jinliu Pan, Yuxi Bi, Yi~Dai, Jiawei Sun, and Haofen Wang. 2023.
\newblock Retrieval-augmented generation for large language models: A survey.
\newblock \emph{arXiv preprint arXiv:2312.10997}.

\bibitem[{Greenhalgh(2004)}]{greenhalgh2004effectiveness}
Trisha Greenhalgh. 2004.
\newblock Effectiveness and efficiency: Random reflections on health services.
\newblock \emph{Bmj}, 328(7438):529.

\bibitem[{Guo et~al.(2024)Guo, Xia, Yu, Ao, and Huang}]{guo2024lightrag}
Zirui Guo, Lianghao Xia, Yanhua Yu, Tu~Ao, and Chao Huang. 2024.
\newblock Lightrag: Simple and fast retrieval-augmented generation.
\newblock \emph{arXiv preprint arXiv:2410.05779}.

\bibitem[{Hu et~al.(2024)Hu, Lei, Zhang, Pan, Ling, and Zhao}]{hu2024grag}
Yuntong Hu, Zhihan Lei, Zheng Zhang, Bo~Pan, Chen Ling, and Liang Zhao. 2024.
\newblock \href {http://arxiv.org/abs/2405.16506} {Grag: Graph retrieval-augmented generation}.

\bibitem[{Jadhav et~al.(2024)Jadhav, Shanbhag, Joshi, Date, and Sonawane}]{jadhav2024maven}
Suramya Jadhav, Abhay Shanbhag, Sumedh Joshi, Atharva Date, and Sheetal Sonawane. 2024.
\newblock Maven at mediqa-corr 2024: Leveraging rag and medical llm for error detection and correction in medical notes.
\newblock In \emph{Proceedings of the 6th Clinical Natural Language Processing Workshop}, pages 374--381.

\bibitem[{Jiang et~al.(2024{\natexlab{a}})Jiang, Fang, Qiu, Zhang, Xu, Chen, Zhang, Zhang, Fang, Chu et~al.}]{jiang2024tc}
Xinke Jiang, Yue Fang, Rihong Qiu, Haoyu Zhang, Yongxin Xu, Hao Chen, Wentao Zhang, Ruizhe Zhang, Yuchen Fang, Xu~Chu, et~al. 2024{\natexlab{a}}.
\newblock Tc-rag: Turing-complete rag's case study on medical llm systems.
\newblock \emph{arXiv preprint arXiv:2408.09199}.

\bibitem[{Jiang et~al.(2023)Jiang, Xu, Gao, Sun, Liu, Dwivedi-Yu, Yang, Callan, and Neubig}]{jiang2023active}
Zhengbao Jiang, Frank~F Xu, Luyu Gao, Zhiqing Sun, Qian Liu, Jane Dwivedi-Yu, Yiming Yang, Jamie Callan, and Graham Neubig. 2023.
\newblock Active retrieval augmented generation.
\newblock \emph{arXiv preprint arXiv:2305.06983}.

\bibitem[{Jiang et~al.(2024{\natexlab{b}})Jiang, Ma, and Chen}]{jiang2024long}
Ziyan Jiang, Xueguang Ma, and Wenhu Chen. 2024{\natexlab{b}}.
\newblock \href {http://arxiv.org/abs/2406.15319} {Longrag: Enhancing retrieval-augmented generation with long-context llms}.

\bibitem[{Jin et~al.(2020)Jin, Pan, Oufattole, Weng, Fang, and Szolovits}]{jin2020medqa}
Di~Jin, Eileen Pan, Nassim Oufattole, Wei-Hung Weng, Hanyi Fang, and Peter Szolovits. 2020.
\newblock \href {http://arxiv.org/abs/2009.13081} {What disease does this patient have? a large-scale open domain question answering dataset from medical exams}.

\bibitem[{Kapoor(2016)}]{kapoor2016types}
Mukul~Chandra Kapoor. 2016.
\newblock Types of studies and research design.
\newblock \emph{Indian journal of anaesthesia}, 60(9):626--630.

\bibitem[{Lewis et~al.(2021)Lewis, Perez, Piktus, Petroni, Karpukhin, Goyal, Küttler, Lewis, tau Yih, Rocktäschel, Riedel, and Kiela}]{lewis2021rag}
Patrick Lewis, Ethan Perez, Aleksandra Piktus, Fabio Petroni, Vladimir Karpukhin, Naman Goyal, Heinrich Küttler, Mike Lewis, Wen tau Yih, Tim Rocktäschel, Sebastian Riedel, and Douwe Kiela. 2021.
\newblock \href {http://arxiv.org/abs/2005.11401} {Retrieval-augmented generation for knowledge-intensive nlp tasks}.

\bibitem[{Li et~al.(2023)Li, Gao, Jiang, Yin, Li, Yan, Zhang, and Yin}]{li2023graph}
Shiyang Li, Yifan Gao, Haoming Jiang, Qingyu Yin, Zheng Li, Xifeng Yan, Chao Zhang, and Bing Yin. 2023.
\newblock Graph reasoning for question answering with triplet retrieval.
\newblock \emph{arXiv preprint arXiv:2305.18742}.

\bibitem[{Masmoudi et~al.(2021)Masmoudi, Lamine, Zghal, Archimede, and Karray}]{masmoudi2021knowledge}
Maroua Masmoudi, Sana Ben Abdallah~Ben Lamine, Hajer~Baazaoui Zghal, Bernard Archimede, and Mohamed~Hedi Karray. 2021.
\newblock Knowledge hypergraph-based approach for data integration and querying: Application to earth observation.
\newblock \emph{Future Generation Computer Systems}, 115:720--740.

\bibitem[{Peng et~al.(2024)Peng, Zhu, Liu, Bo, Shi, Hong, Zhang, and Tang}]{peng2024graph}
Boci Peng, Yun Zhu, Yongchao Liu, Xiaohe Bo, Haizhou Shi, Chuntao Hong, Yan Zhang, and Siliang Tang. 2024.
\newblock Graph retrieval-augmented generation: A survey.
\newblock \emph{arXiv preprint arXiv:2408.08921}.

\bibitem[{Procko(2024)}]{procko2024graph}
Tyler Procko. 2024.
\newblock Graph retrieval-augmented generation for large language models: A survey.
\newblock \emph{Available at SSRN}.

\bibitem[{Ram et~al.(2023)Ram, Levine, Dalmedigos, Muhlgay, Shashua, Leyton-Brown, and Shoham}]{ram2023context}
Ori Ram, Yoav Levine, Itay Dalmedigos, Dor Muhlgay, Amnon Shashua, Kevin Leyton-Brown, and Yoav Shoham. 2023.
\newblock In-context retrieval-augmented language models.
\newblock \emph{Transactions of the Association for Computational Linguistics}, 11:1316--1331.

\bibitem[{Reimers and Gurevych(2019)}]{reimers-2019-sentence-bert}
Nils Reimers and Iryna Gurevych. 2019.
\newblock \href {https://arxiv.org/abs/1908.10084} {Sentence-bert: Sentence embeddings using siamese bert-networks}.
\newblock In \emph{Proceedings of the 2019 Conference on Empirical Methods in Natural Language Processing}. Association for Computational Linguistics.

\bibitem[{Sarmah et~al.(2024)Sarmah, Mehta, Hall, Rao, Patel, and Pasquali}]{sarmah2024hybridrag}
Bhaskarjit Sarmah, Dhagash Mehta, Benika Hall, Rohan Rao, Sunil Patel, and Stefano Pasquali. 2024.
\newblock Hybridrag: Integrating knowledge graphs and vector retrieval augmented generation for efficient information extraction.
\newblock In \emph{Proceedings of the 5th ACM International Conference on AI in Finance}, pages 608--616.

\bibitem[{Shao et~al.(2023)Shao, Gong, Shen, Huang, Duan, and Chen}]{shao2023enhancing}
Zhihong Shao, Yeyun Gong, Yelong Shen, Minlie Huang, Nan Duan, and Weizhu Chen. 2023.
\newblock Enhancing retrieval-augmented large language models with iterative retrieval-generation synergy.
\newblock \emph{arXiv preprint arXiv:2305.15294}.

\bibitem[{Team(2024)}]{qwen2.5}
Qwen Team. 2024.
\newblock \href {https://qwenlm.github.io/blog/qwen2.5/} {Qwen2.5: A party of foundation models}.

\bibitem[{Tonmoy et~al.(2024)Tonmoy, Zaman, Jain, Rani, Rawte, Chadha, and Das}]{tonmoy2024comprehensive}
SM~Tonmoy, SM~Zaman, Vinija Jain, Anku Rani, Vipula Rawte, Aman Chadha, and Amitava Das. 2024.
\newblock A comprehensive survey of hallucination mitigation techniques in large language models.
\newblock \emph{arXiv preprint arXiv:2401.01313}.

\bibitem[{Wang et~al.(2024{\natexlab{a}})Wang, Li, Jiang, Tian, Wang, Luo, Tang, Cheng, Zhao, and Gao}]{wang2024blendfilter}
Haoyu Wang, Ruirui Li, Haoming Jiang, Jinjin Tian, Zhengyang Wang, Chen Luo, Xianfeng Tang, Monica Cheng, Tuo Zhao, and Jing Gao. 2024{\natexlab{a}}.
\newblock Blendfilter: Advancing retrieval-augmented large language models via query generation blending and knowledge filtering.
\newblock \emph{arXiv preprint arXiv:2402.11129}.

\bibitem[{Wang et~al.(2024{\natexlab{b}})Wang, Wang, Gao, Zhang, Wu, Xu, Shi, Wang, Li, Qian et~al.}]{wang2024searching}
Xiaohua Wang, Zhenghua Wang, Xuan Gao, Feiran Zhang, Yixin Wu, Zhibo Xu, Tianyuan Shi, Zhengyuan Wang, Shizheng Li, Qi~Qian, et~al. 2024{\natexlab{b}}.
\newblock Searching for best practices in retrieval-augmented generation.
\newblock In \emph{Proceedings of the 2024 Conference on Empirical Methods in Natural Language Processing}, pages 17716--17736.

\bibitem[{Wang et~al.(2023)Wang, Araki, Jiang, Parvez, and Neubig}]{wang2023learning}
Zhiruo Wang, Jun Araki, Zhengbao Jiang, Md~Rizwan Parvez, and Graham Neubig. 2023.
\newblock Learning to filter context for retrieval-augmented generation.
\newblock \emph{arXiv preprint arXiv:2311.08377}.

\bibitem[{Wu et~al.(2024)Wu, Zhu, Qi, Chen, Xu, Menolascina, and Grau}]{wu2024medical}
Junde Wu, Jiayuan Zhu, Yunli Qi, Jingkun Chen, Min Xu, Filippo Menolascina, and Vicente Grau. 2024.
\newblock Medical graph rag: Towards safe medical large language model via graph retrieval-augmented generation.
\newblock \emph{arXiv preprint arXiv:2408.04187}.

\bibitem[{Xidong~Wang(2023)}]{cmedbenchmark}
Dingjie~Song Xidong~Wang, Guiming Hardy~Chen. 2023.
\newblock Cmb: Chinese medical benchmark.
\newblock \url{https://github.com/FreedomIntelligence/CMB}.
\newblock Xidong Wang, Guiming Hardy Chen, Dingjie Song, and Zhiyi Zhang contributed equally to this github repo.

\bibitem[{Xu et~al.(2024{\natexlab{a}})Xu, Qi, Guo, Wang, Wang, Zhang, and Xu}]{xu2024knowledgeconflictsllmssurvey}
Rongwu Xu, Zehan Qi, Zhijiang Guo, Cunxiang Wang, Hongru Wang, Yue Zhang, and Wei Xu. 2024{\natexlab{a}}.
\newblock \href {http://arxiv.org/abs/2403.08319} {Knowledge conflicts for llms: A survey}.

\bibitem[{Xu et~al.(2024{\natexlab{b}})Xu, Cruz, Guevara, Wang, Deshpande, Wang, and Li}]{xu2024retrieval}
Zhentao Xu, Mark~Jerome Cruz, Matthew Guevara, Tie Wang, Manasi Deshpande, Xiaofeng Wang, and Zheng Li. 2024{\natexlab{b}}.
\newblock Retrieval-augmented generation with knowledge graphs for customer service question answering.
\newblock In \emph{Proceedings of the 47th International ACM SIGIR Conference on Research and Development in Information Retrieval}, pages 2905--2909.

\bibitem[{Yan et~al.(2024)Yan, Gu, Zhu, and Ling}]{yan2024corrective}
Shi-Qi Yan, Jia-Chen Gu, Yun Zhu, and Zhen-Hua Ling. 2024.
\newblock Corrective retrieval augmented generation.
\newblock \emph{arXiv preprint arXiv:2401.15884}.

\bibitem[{Yang et~al.(2024)Yang, Yang, Feng, Ouyang, Blum, She, Jiang, Lecue, Lu, and Li}]{yang2024graphusion}
Rui Yang, Boming Yang, Aosong Feng, Sixun Ouyang, Moritz Blum, Tianwei She, Yuang Jiang, Freddy Lecue, Jinghui Lu, and Irene Li. 2024.
\newblock Graphusion: A rag framework for knowledge graph construction with a global perspective.
\newblock \emph{arXiv preprint arXiv:2410.17600}.

\bibitem[{Ye et~al.(2023)Ye, Liu, Zhang, Hua, and Jia}]{ye2023cognitive}
Hongbin Ye, Tong Liu, Aijia Zhang, Wei Hua, and Weiqiang Jia. 2023.
\newblock Cognitive mirage: A review of hallucinations in large language models.
\newblock \emph{arXiv preprint arXiv:2309.06794}.

\bibitem[{Yu et~al.(2024)Yu, Ping, Liu, Wang, You, Zhang, Shoeybi, and Catanzaro}]{yu2024rank}
Yue Yu, Wei Ping, Zihan Liu, Boxin Wang, Jiaxuan You, Chao Zhang, Mohammad Shoeybi, and Bryan Catanzaro. 2024.
\newblock \href {http://arxiv.org/abs/2407.02485} {Rankrag: Unifying context ranking with retrieval-augmented generation in llms}.

\end{thebibliography}
\bibliographystyle{acl_natbib}

\appendix
\begin{figure*}
    \centering
    \includegraphics[width=1.0\linewidth]{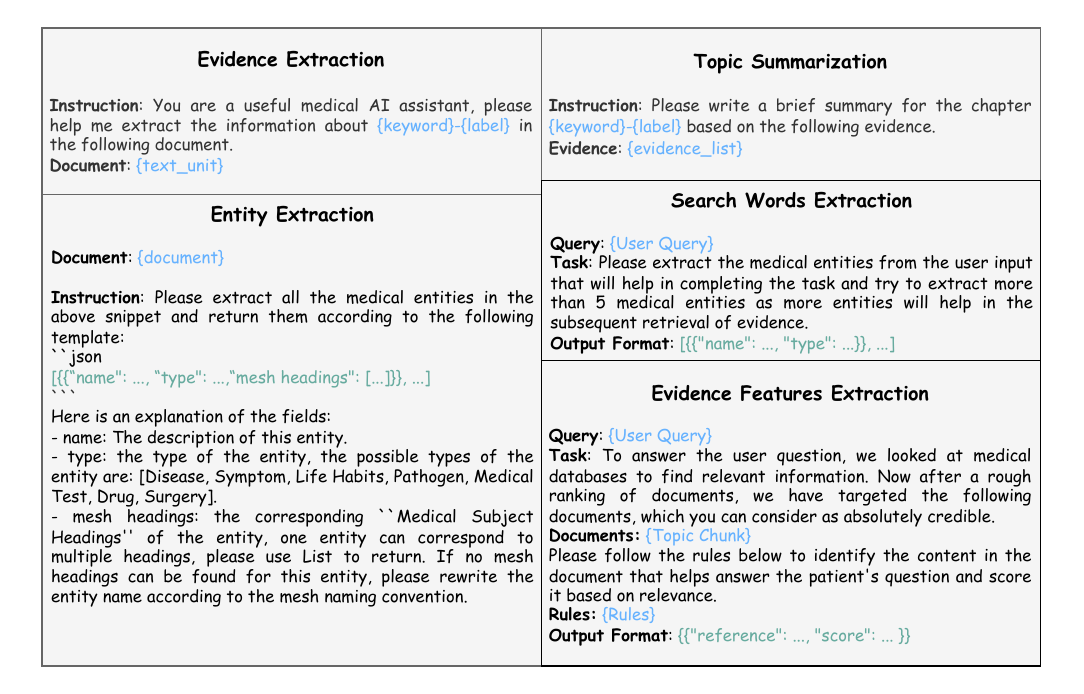}
    \caption{Prompts used by Graph Construction and Evidence Retrieval.}
    \label{fig:all_prompt}
\end{figure*}

\section{Appendix}
\label{sec:appendix}

\subsection{Hyper-Relation Mapping}
\label{sec:label_def}
To facilitate the extraction of evidence from medical literature, it is necessary to identify the relevant keywords and pre-define the corresponding hyper-relations. Given that our primary sources are drug instructions and disease diagnosis and treatment guidelines, we will concentrate on two types of entity: drugs and diseases. Other types of entities will be addressed in the subsequent formal entity extraction process. The hyper-relation mappings are as follows.
\begin{itemize}
\item Diseases: symptoms, causes, diagnosis, treatment, and prognosis.
\item Drugs: treatment, usage, adverse reactions, contraindications, drug interactions, and precautions.
\end{itemize}

\subsection{Prompts Used in Our Work}
\label{sec:prompt}
Figure~\ref{fig:all_prompt} displays the prompts used in our study, consisting of five distinct cue templates. In the following, we provide a detailed description of each template.

\paragraph{Evidence Extraction} 
a document refers to one of our segmented text units. The length of these units is determined by the window size; a smaller window size yields more detailed evidence but increases the cost of graph construction.

\paragraph{Entity Extraction}
we standardize entity names by directing the LLM to output Medical Subject Headings (MeSH) concurrently as it recognizes entities. MeSH provides a consistent framework for medical terms widely used throughout the healthcare industry. An individual entity may correspond to multiple MeSH terms, enabling its association with various categories. In such instances, we decompose the entity into several standard MeSH terms to expand the graph vocabulary. For EbmKG, we keep only the standardized entity names and eliminate the original ones.

\paragraph{Topic Summarization}
We employ LLM to condense evidence with identical labels. In practice, numerous topics contain excessive evidence, hindering LLM from summarizing all at once. In these instances, we initially summarize evidence in batches for local summaries, and then LLM produces final summaries from these local summaries.

\paragraph{Search Words Extraction} We direct the LLM to produce at least five search terms for each query. Generating a larger number of terms can significantly improve the effectiveness of future search.

\begin{figure*}[thbp!]
    \centering
    \includegraphics[width=1.0\linewidth]{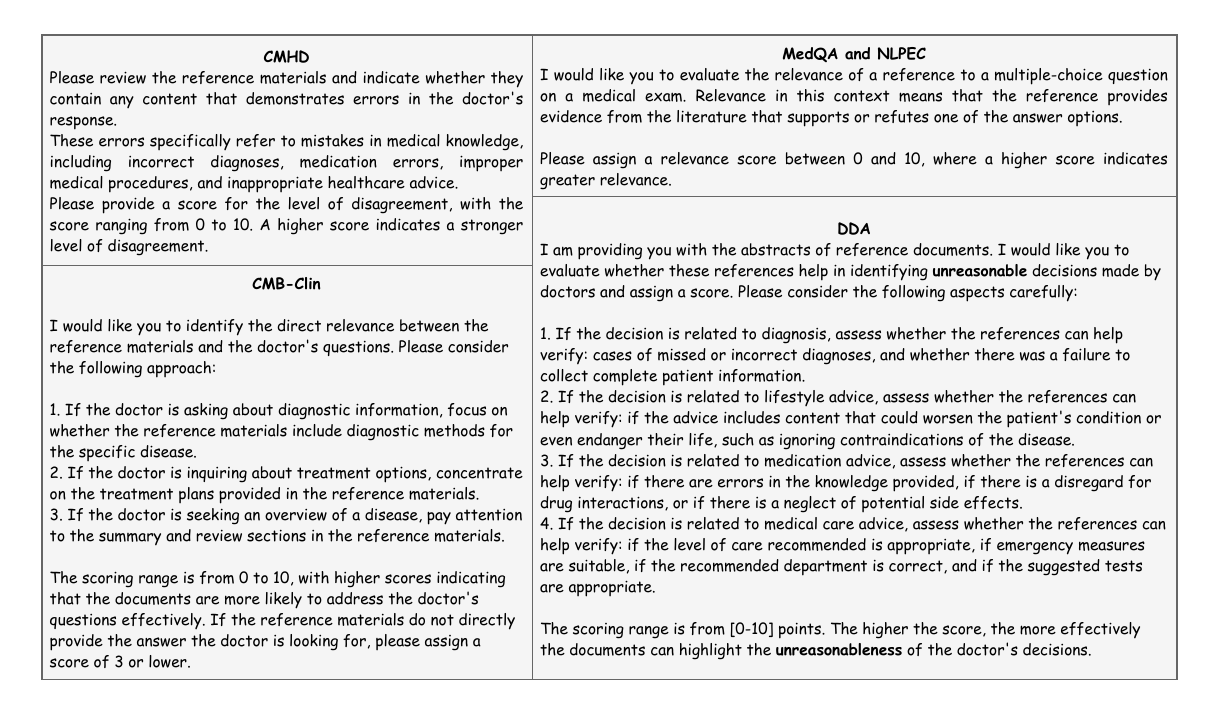}
    \caption{Search conditions for different datasets}
    \label{fig:score_rules}
\end{figure*}

\paragraph{Evidence Features Extraction}
We allow users to adjust the significance of evidence via the "Rules" slot (search condition), enhancing IDEP's capability to fulfill the varied demands of different tasks with these extra query conditions. The specific query conditions applied in this study are detailed in Figure~\ref{fig:score_rules}, we build specialized search condition for each type of task.

\subsection{Implementation Details}
\label{sec:imp_details}
The knowledge base, derived from 41,504 documents, comprises 11,024 drug descriptions sourced from the Chinese People's Health website\footnote{\url{https://pharmacy.pmphai.com/}} and the remainder from medical guidelines on the UpToDate site\footnote{\url{https://www.uptodate.cn/home}}, with 10,000 in English and the rest in Chinese. These documents were randomly crawled, and we did not require that the answers to the questions in the test set be explicitly found within the documents. This aRandom Walkoach enhances the practical application significance of our work.

A consistent slicing window of 1,024 tokens with a 200-token overlap was applied. The construction of the knowledge base involved the use of the language model \textit{Qwen2.5-72B-instruct} and the text embedding model \textit{text-ada-embedding-002}\footnote{\url{https://openai.com/}}. For GraphRAG and EbmKG, the process utilized 16 A100-80G GPUs over a period of two weeks. During testing, the retrieved documents were limited to 8,000 words to remain within the context capacity of the LLM. The LLM sampling temperature was set to 0 during generation.

The following statistics summarize the constructed knowledge base:

\begin{itemize}
    \item \textbf{VectorRAG:} 1,580,263 document segments were generated.
    \item \textbf{GraphRAG:} This consists of 1,216,699 entities, 3,004,725 relationships, and 159,586 communities, organized into seven levels.
    \item \textbf{RAG4EBM:} This includes 251,849 entities, 433,611 topics (hyperedges), and 806,495 evidence pieces. The hyperedge star pattern extends to a total of 11,578,906 edges.
\end{itemize}

\paragraph{Token Usage}
\begin{figure}
    \centering
    \includegraphics[width=1.0\linewidth]{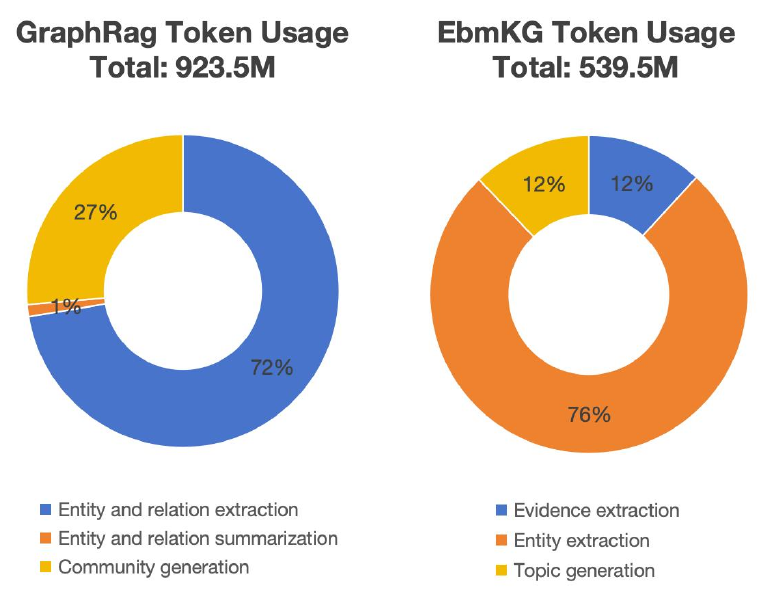}
    \caption{Number of Token consumed by GraphRAG and EbmKG constructs.}
    \label{fig:token_usage}
\end{figure}

LLMs' high computational costs hinder practical knowledge graph construction. Our aRandom Walkoach, EbmKG, addresses this by reducing overhead. As shown in Figure~\ref{fig:token_usage}, EbmKG achieves 41\% lower overhead than GraphRAG, highlighting its superior resource efficiency.

We observe that the primary computational cost for both methods lies in the extraction of entities and relations. Consequently, to further reduce computational costs, the implementation of a small Named Entity Recognition (NER) model, as opposed to a LLM, presents a viable strategy. However, this aRandom Walkoach is not entirely applicable to GraphRAG, which need to extract the entity and relation descriptions. In contrast to GraphRAG, our aRandom Walkoach does not necessitate entity descriptions, thereby affording greater potential for optimization. The implementation of a smaller NER model in EbmKG would reduce EbmKG construction costs to 14\% of those incurred by GraphRAG.

Furthermore, the incorporation of new documents into GraphRAG entails a complex process, necessitating the recalculation of all community structures, which accounts for 27\% of the overall computational costs. EbmKG, conversely, employs keyword tags for independent grouping, thereby affecting only topics that are closely related. This characteristic enables EbmKG to accommodate document updates with minimal disruption, thereby exceeding GraphRAG in terms of both flexibility and efficiency.

\subsection{Detailed Dataset Descriptions}
\label{sec:data_des}
In this section, we describe in detail the testing format for each dataset.
The statistics of the test sets we used are shown in Table~\ref{tab:test_sample}.

\begin{table}
\centering
\begin{tabular}{@{}lcc@{}}
\toprule
Testset                      & Sample Num \\ \midrule
MedQA (Mainland)      & 3426       \\
NLPEC                 & 550        \\
CMHE-D           & 2000       \\
CMHE-G           & 994        \\
DDA              & 100        \\
CMB-Clin         & 202        \\ \bottomrule
\end{tabular}
\caption{Test Sample Statistics}
\label{tab:test_sample}
\end{table}

\paragraph{Multi-choice Q\&A}

\begin{figure*}[htbp]
    \centering
    \includegraphics[width=1.0\linewidth]{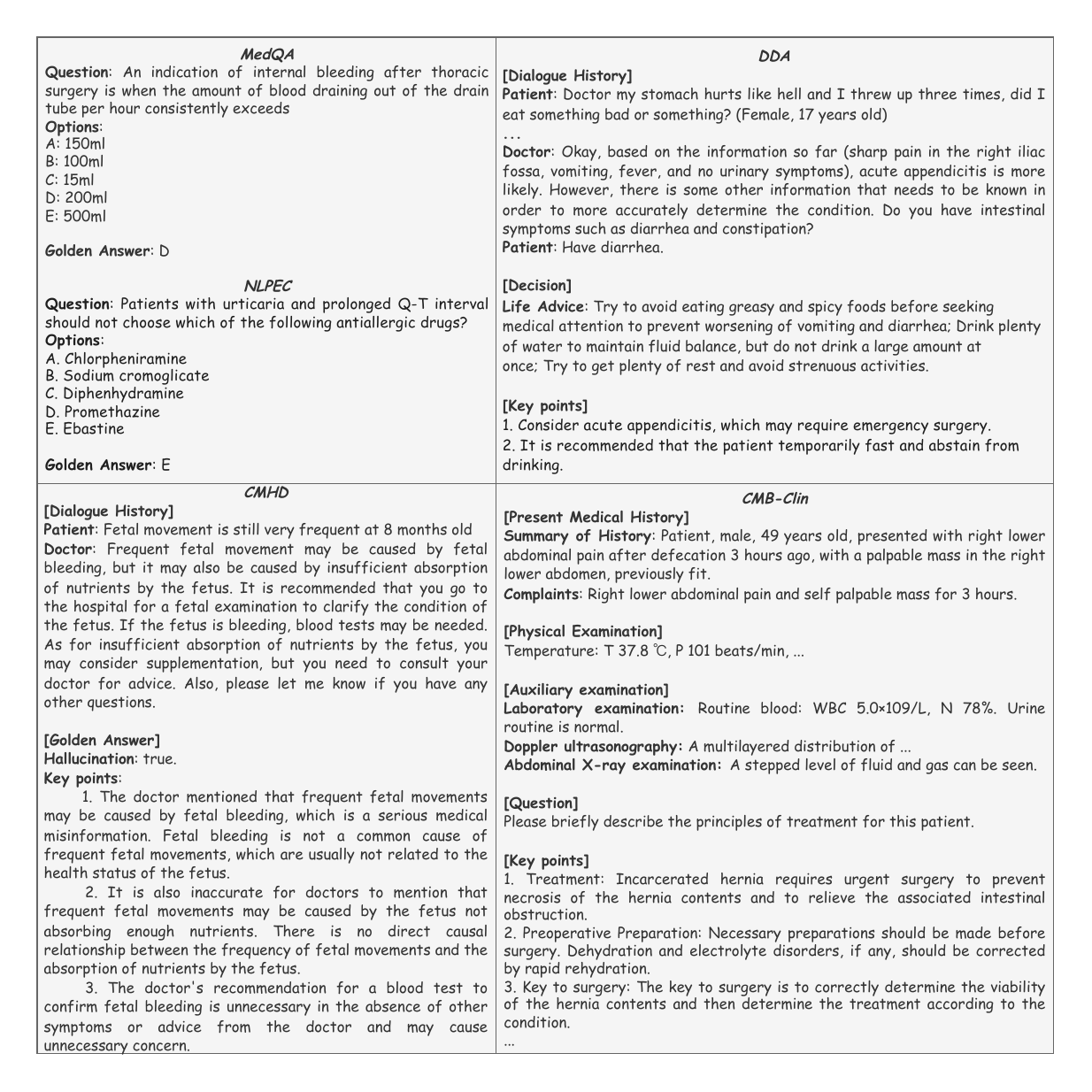}
    \caption{Samples of Test Sets}
    \label{fig:dataset}
\end{figure*}

commonly used to assess LLMs incorporates answer options, providing a clear target for responses. We propose MCQA evaluates the thoroughness and accuracy of an LLM's knowledge. Our aRandom Walkoach uses two datasets: MedQA and NLPEC, as shown in Figure~\ref{fig:dataset} (MedQA, NLPEC). MedQA questions are related to medical licensing exams that cover clinical, biomedical, and policy topics, while NLPEC relates to pharmacy licensing and drug application. Since our knowledge base is derived from drug inserts and medical guidelines, many MedQA questions remain unanswerable from our materials, unlike NLPEC questions, which are easily addressed. We suggest this setup aptly demonstrates the performance of various RAG methods with out-of-distribution queries.

\paragraph{CMHD} was utilized to test the model's ability to handle hallucinations. We classified these data into CHMD-D and CHMD-G: 1) CHMD-D measures the LLM's skill in identifying hallucinations. 2) CHMD-G requires the model to identify the hallucination locations. Examples of these tasks are shown in Figure~\ref{fig:dataset}(CMHD). For CHMD-D, we prompt the LLM to determine if the doctor's reply has an illusion, labeling it True or False. For CHMD-G, we ask the LLM to create a text explaining the illusion in the doctor's response. We use the explanation coverage of key points as a metric.

\paragraph{DDA} is designed to assess intricate medical decisions. We utilized GPT-4 for simulated doctor-patient dialogs to derive medical decisions, subsequently reviewed by seasoned physicians for any issues. The physicians' recommendations serve as standard answers for assessment. Our objective is to evaluate the model's competence in diagnosis, treatment planning, lifestyle advice, and medical guidance. Figure~\ref{fig:dataset}(DDA) illustrates the DDA evaluation format.

\paragraph{CMB-clin} requires the model to interpret a patient's examination report to generate responses to specific questions. Since the original dataset provides reference answers in a nonformalized text format, it is challenging to use it directly for generating quantitative metrics. Therefore, we manually decomposed the raw standardized answers into key points, as illustrated in Figure~\ref{fig:dataset} (CMB-Clin) and evaluated them using the same methodology as DDA and CHMD-G.

\begin{figure}
    \centering
    \includegraphics[width=1.0\linewidth]{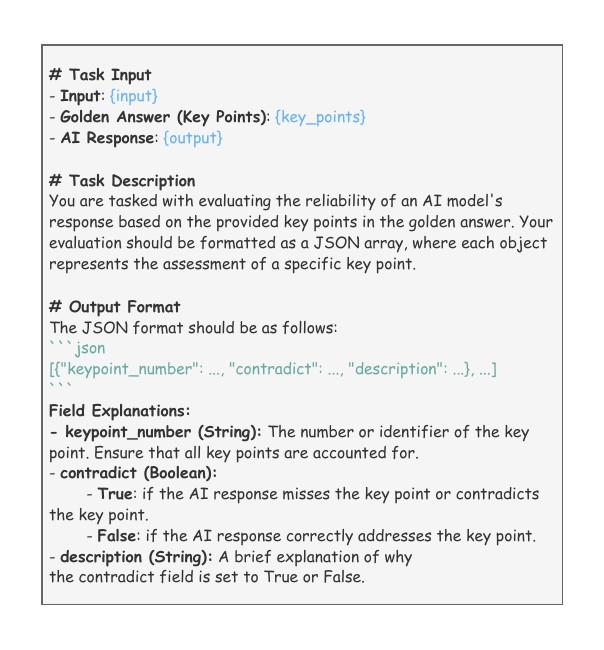}
    \caption{The prompt of key points evaluation}
    \label{fig:kp}
\end{figure}

\subsection{Key Points Evaluation}
\label{sec:kp_eval}

Given that conventional metrics such as BLEU and ROUGE struggle to effectively assess the quality of LLM output, we introduce an evaluation method based on key points. Our aRandom Walkoach begins by manually breaking down the reference answers for each generated task into distinct keypoints. Following this, GPT-4o is tasked with assessing how well the LLM responses cover these keypoints, the prompt is illustrated in Figure~\ref{fig:kp}. Formally, we define the test set as $D = <Q, K>$, where $Q$ represents the question and $K$ represents the keypoint related to the response to the question. The LLM score is computed as follows:
\begin{equation}
    \text{Score} = \frac{\sum_i^n |\{k \mid k \in K_i, \text{not contradict}(k)\}|}{\sum_i^n |K_i|}
\end{equation}

\subsection{Metrics for Retrieval Evaluation }
\label{sec:appendix_com}

\begin{figure*}
    \centering
    \includegraphics[width=1.0\linewidth]{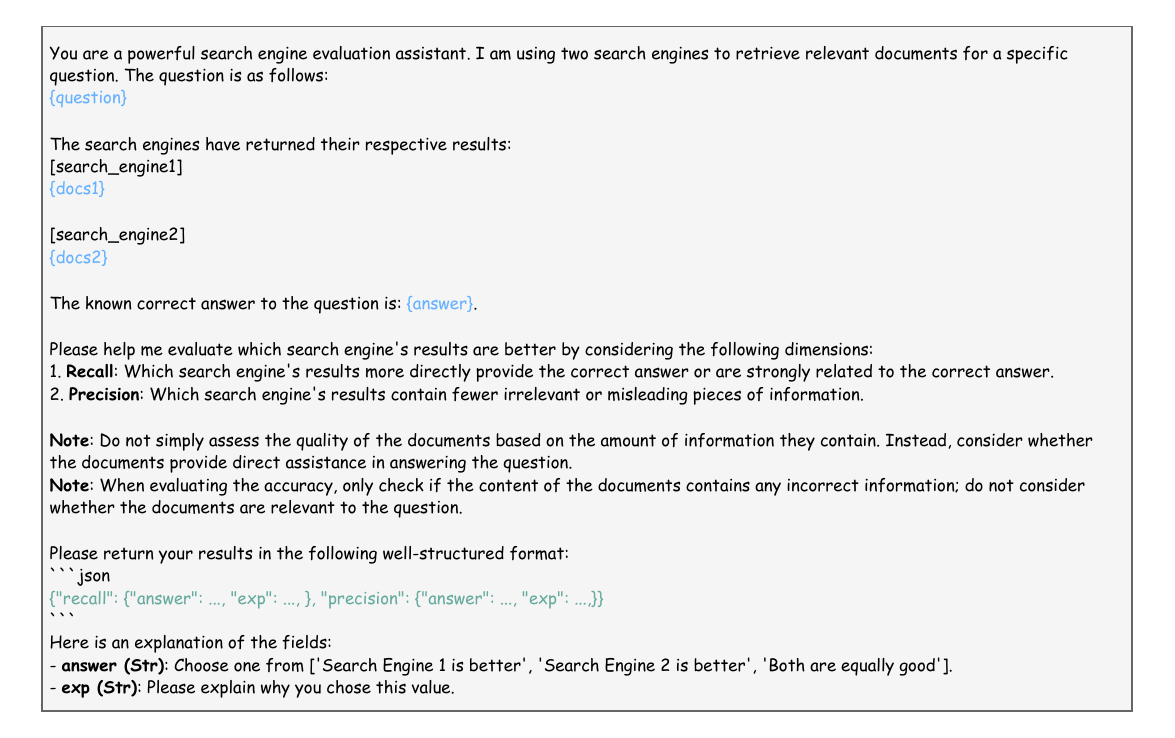}
    \caption{Prompt for calculating Recall and Precision.}
    \label{fig:compaire}
\end{figure*}

To assess document quality, we employed recall and precision as our primary criteria. Recall measures how directly search engine results match the standard answer, while precision assesses the degree of irrelevant or misleading content. These scores were generated by guiding GPT-4o through the Prompt (see Figure~\ref{fig:compaire}). We then introduce the Advantage Score, a combination of these metrics, as follows:
\begin{gather}
r = \frac{r_{win}}{1 - r_{tie}}, \quad p = \frac{p_{win}}{1 - p_{tie}} \\
a = \frac{4rp}{(r + p)} - 1
\end{gather}
In this formula, $a$ denotes the Advantage Score, with $r$ and $p$ indicating RAG4EBM's winning proportions in Recall and Precision, respectively. A positive $a$ suggests that EbmKG retrieves more relevant documents.

\subsection{Ablation Study}
\label{sec:ab_study}
\begin{figure}[htbp]
    \centering
    \includegraphics[width=1.0\linewidth]{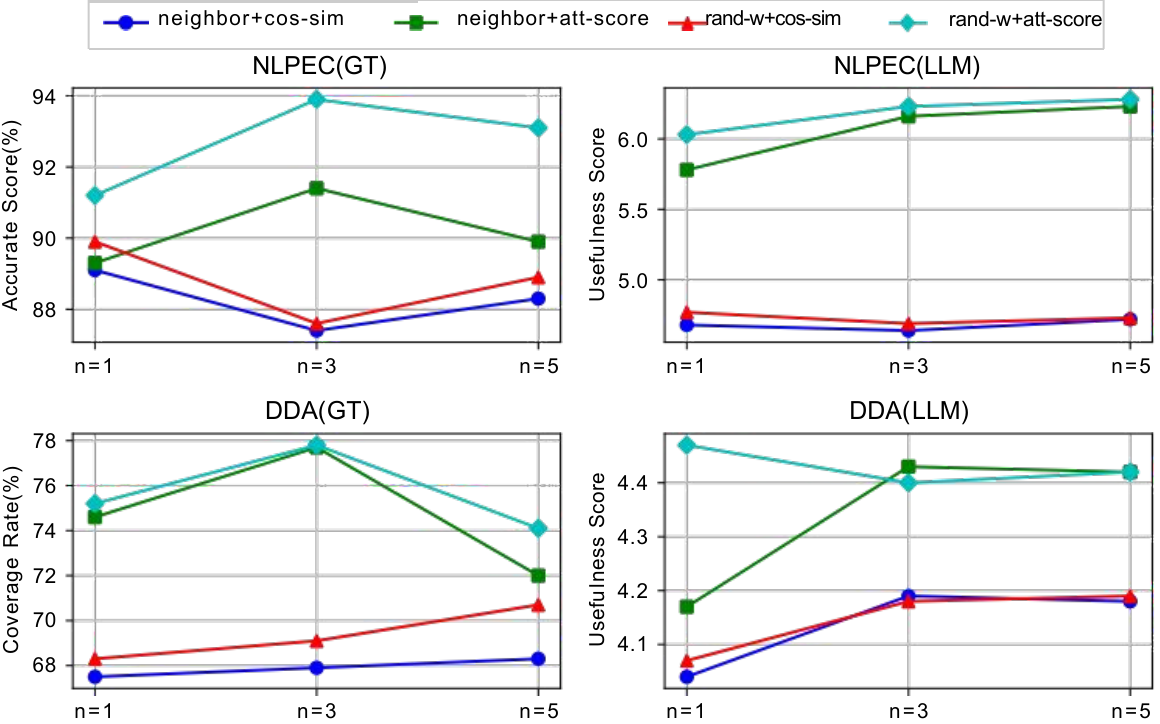}
    \caption{Performances of our proposed method with different components on dataset NLPEC and DDA, where the horizontal coordinate indicates the number of entities linked to each keyword ($n$) and the vertical coordinates denotes two evaluations methods with ground truth(`GT') and LLM grading (`LLM').}
    \label{fig:ab_study}
\end{figure}

\begin{figure}
    \centering
    \includegraphics[width=1.0\linewidth]{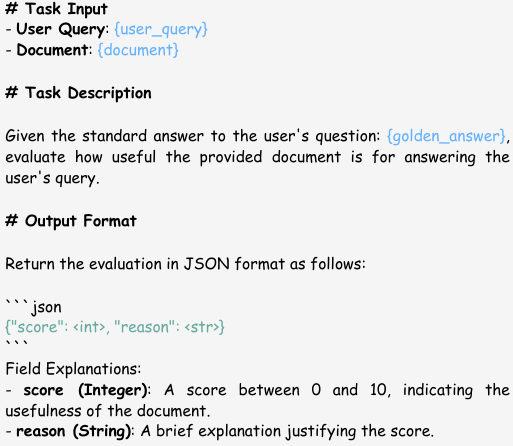}
    \caption{The prompt of Usefulness Evaluation}
    \label{fig:usefull}
\end{figure}

We conducted ablation studies on the NLPEC and DDA datasets to identify key factors affecting RAG4EBM performance, using 100 randomly selected samples with varying query complexities. Figure~\ref{fig:ab_study} shows two evaluation methods: 1) GT Evaluation, comparing datasets against ground truth answers, and 2) LLM Evaluation, rating GPT-4o output documents on a 0-10 scale for utility, the prompt we used is shown in Figure~\ref{fig:usefull}. We examined three parameters: the number of entities linked per keyword ($n$), topic sampling approach (neighbor sampling vs. random walk), and evidence ranking strategy (vector cosine similarity vs. attention score).

\paragraph{Impact of Linked Entities per Keyword ($n$).} As $n$ increases from 1 to 3, there is significant performance variability across all RAG4EBM models for both datasets and evaluation metrics (3 on GT and 0.25 on LLM evaluation). However, from $n=3$ to $n=5$, these changes are less pronounced, especially for LLM evaluation. In the DDA dataset with GT evaluation, an increase in $n$ causes a clearer performance drop, likely due to the retrieval of more irrelevant documents, which adds noise.

\paragraph{Neighbor Sampling vs. Random Walk.} A comparison between `neighbor+cos-sim' and `random walk+cos-sim' shows that Random Walk consistently outperforms neighbor sampling. This holds true when the ranking metric is switched from `cos-sim' to `attention score', suggesting that the improvement of Random Walk in topic sampling is modest but consistent.

\paragraph{Vector cosine similarity vs. attention score.} Comparing `neighbor+cos-sim' with `neighbor+attention score' shows that attention score significantly enhances performance in all scenarios, achieving improvements up to 10 on GT and 1.5 on LLM. This is consistent with the Random Walk method. The results clearly exhibit the attention score method's superiority in accurately searching evidence with complex queries.

In conclusion, evidence ranking methods have the most significant effect, followed by the selection of topic sampling strategies, whereas the number of linked entities influences the results the least. For the usefulness score, the impact of the IDEP is the most significant. In addition, the benefit of Random Walk is evident in the GT Evaluation. In addition, we also find that IDEP has a higher gain than NLPEC on the DDA dataset, which suggests that the scheme is more effective for dealing with complex problems. These findings provide strong evidence for the effectiveness of our proposed method.

\subsection{Case Study of GraphRAG}
\label{sec:case_study}

\begin{figure*}[thbp!]
    \centering
    \includegraphics[width=1.0\linewidth]{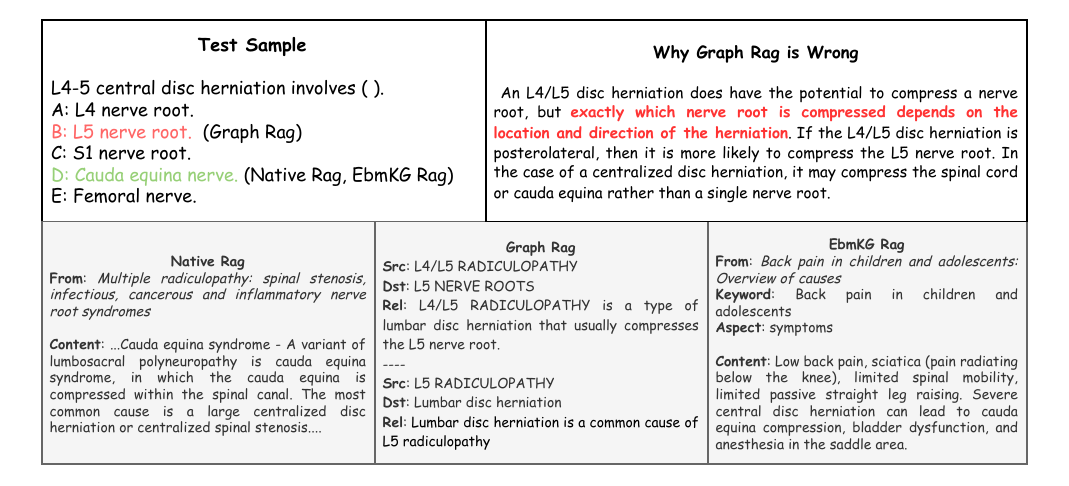}
    \caption{Comparisons of results generated by various RAG aRandom Walkoaches.}
    \label{fig:case_study}
\end{figure*}

In our case study, we randomly picked 200 samples from the MedQA dataset. GraphRAG made errors in 21 of these cases. Among these 21 errors, in 14 instances, GPT-4o judged GraphRAG's search results to be superior to those from VectorRAG and RAG4EBM. Manual review showed that while the literature retrieved by GraphRAG was highly relevant to the test samples, all 14 contained substantially misleading information undetected by current LLMs. Figure~\ref{fig:case_study} shows an example.

In this instance, GraphRAG incorrectly extracted the statement ``L4 / L5 RADICULOPATHY is a type of lumbar disc herniation that usually compresses the L5 nerve root.'' The error occurs because, during the knowledge graph extraction via LLM, the complex relationship is overly broken down, as demonstrated in Figure~\ref{fig:intro}, losing the conditional variable of the L4/L5 RADICULOPATHY direction. In contrast, VectorRAG and EbmKG RAG do not face this problem and accurately identify documents pertinent to the test sample.

\subsection{Error Type Definition}
\label{sec:etd}
In this section, we provide a detailed definition of the error types. We have categorized these errors into three main types: Lack of Relevant Documents, LLM Inference Errors, and Suboptimal Evidence Retrieval, based on their underlying causes. The following subsections will elaborate on each of these error types.

\begin{figure*}[thbp!]
    \centering
    \includegraphics[width=1.0\linewidth]{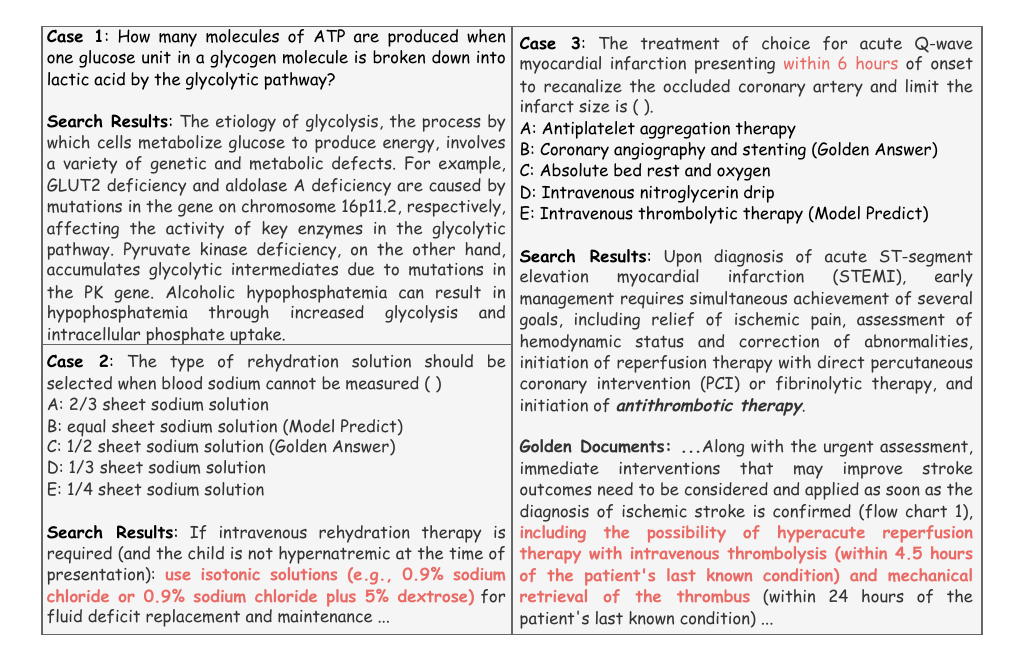}
    \caption{Examples of different error types.}
    \label{fig:etp}
\end{figure*}

\paragraph{Lack of Relevant Documents}
Since the documents we selected were mainly from clinical medical guidelines and drug descriptions. When responding to questions related to, for example, biochemistry, it was not possible to find aRandom Walkopriate documents, an example of which is shown in Figure~\ref{fig:etp} Case 1. In this example, the evidence we were able to retrieve at best covered treatment protocols for glucose-related diseases, and could not answer questions related to glucose chemistry.

\paragraph{LLM Inference Errors}
Even if IdepRAG successfully retrieves relevant literature, the language model (LLM) may still fail to effectively utilize this information. This is illustrated in Figure~\ref{fig:etp}, case 2. The retrieved evidence clearly supports the use of an isotonic solution, specifically a 1/2 sheet sodium solution, as the correct answer. However, the model incorrectly chose an isotonic sodium solution. This discrepancy indicates that the model is unable to correctly correlate the term "isotonic solution" with "1/2 sheet sodium solution."

\paragraph{Suboptimal Evidence Retrieval}
IdepRAG continues to struggle with the granular perception of information during evidence retrieval, which can result in suboptimal query outcomes. As illustrated in Figur~\ref{fig:etp}, Case 3, while IdepRAG successfully retrieved evidence that intravenous thrombolytic therapy is effective for treating Q-wave myocardial infarction, it failed to capture a critical detail: the therapy is only indicated for patients within 4.5 hours of symptom onset.

\end{document}